\documentclass[11pt]{article}
\usepackage[margin=1in]{geometry}
\usepackage[utf8]{inputenc}
\usepackage[T1]{fontenc}
\usepackage{amsmath,amssymb}
\usepackage{booktabs}      
\usepackage{graphicx}
\graphicspath{{figures/}}  
\usepackage{hyperref}
\usepackage{xcolor}
\usepackage[numbers]{natbib}
\usepackage{url}           

\newcommand{\rrho}{\rho}
\newcommand{\wdist}{\mathrm{WD}}
\newcommand{\starhead}{L11\,H16}

\title{Readable, Faithful, Used:\\
Three Dissociable Properties of Demographic Identity\\
in a Language Model}

\author{Fathin Difa Robbani\\
Independent Researcher\\
\small ORCID: \href{https://orcid.org/0009-0000-6184-8919}{0009-0000-6184-8919}}

\date{}

\begin{document}
\maketitle

\begin{abstract}
Large language models are widely used to simulate survey respondents, yet
their outputs are homogeneous and unfaithful to real inter-group
differences, and whether this reflects what a model knows or uses has
remained untested. Using representational similarity analysis against Pew
American Trends Panel ground truth, we score demographic read-out
locations in Mistral-7B and intervene causally across six attribute
types. The internal geometry is faithful: attention-head read-outs
dominate the standard residual read-out, reaching selection-corrected
$\rrho$ up to $0.63$---about $70\%$ of the measurement-reliability
ceiling---and one head, \starhead, is significantly faithful across all
six types, though race-based types stay weak and prompt-fragile,
replicating in a second model family. Yet causal use does not track
fidelity: the clearest causal pathway ($p=0.002$) sits in one of the
least faithful types, the most faithful type shows no
correction-surviving effect, and full identity swaps in the prompt move
predictions by under $2\%$ of their error. A $128$-dimensional probe on
that head lands $21$--$31\%$ closer to survey truth than the model's
answers, yet recovers almost none of the per-question group ordering.
Readable, faithfully arranged, and causally used are three dissociable
properties of the same model; treating them as one claim is what keeps
the ``can LLMs simulate populations'' debate unresolved.
\end{abstract}

\section{Introduction}
\label{sec:intro}

LLMs are increasingly deployed as simulated survey respondents and human
simulators \citep{subpop2025,odyssim2026}. Yet behavioral evidence shows
their outputs are excessively uniform: simulated users are ``overly
agreeable'' and ``stylistically uniform'' \citep{sim2real2026}, opinion
predictions barely differentiate demographic groups, and general model
capability does not predict simulation fidelity \citep{sim2real2026}. This
evidence is \emph{behavioral}---measured from the outside. Whether the
failure lives in what the model \emph{knows} or in what it \emph{uses} has
remained open, because the two research communities each hold only half
of the required instrumentation.

Interpretability work localizes demographic and persona information inside
models---down to individual attention heads \citep{llmopinions2026,
kimevans2025}---but validates against the model's own behavior (probe
accuracy, steering success), never against external population ground
truth. LLM-survey-simulation work has exactly that ground truth (real
per-group response distributions) but treats the model as a black box.
The experiment that requires both instruments at once---whether the
internal arrangement of groups matches how those groups actually
answer---sits in the gap between them and had not been run.

We hold both instruments at once. For 169 intersectional demographic
cells (e.g., \textit{Asian $\times$ Hindu}, \textit{age 18--29 $\times$
Democrat}) we compute two representational dissimilarity matrices---one
from real Pew ATP response distributions (Wasserstein distances between
groups' answers), one from the model's internal representations at a
given location---and correlate them (RSA; \citealp{kriegeskorte2008}).
Applied at 1{,}089 locations $\times$ 4 prompt operationalizations, this
yields a \emph{fidelity map}: where inside the model does the arrangement
of demographic groups mirror how the groups actually differ in opinion?

Our results split the question ``can LLMs simulate populations?'' into
three separable properties (we reserve ``layer'' for transformer layers):
\begin{enumerate}
  \item \textbf{Readable} (established by prior work): per-group opinion
        distributions can be decoded from internal activations
        \citep{llmopinions2026}.
  \item \textbf{Faithfully arranged} (this work): at specific locations,
        including a single attention head, \starhead, the \emph{inter-group
        geometry} mirrors real opinion structure (selection-corrected
        $\rrho$ up to $0.63$), which per-group readability neither implies
        nor requires.
  \item \textbf{Actually used} (this work): the faithfully arranged
        information barely flows into the model's answers---full identity
        swaps in the prompt move predictions by ${<}2\%$ of their error,
        and not directionally toward the target group's truth. A
        strong-instrument intervention does find a real causal pathway,
        but where it is clearest is \emph{not} where the map is most
        faithful (Sec.~\ref{sec:sweep}).
\end{enumerate}

\paragraph{Contributions.}
\begin{itemize}
  \item A \emph{fidelity map} of demographic identity in an LLM---to our
        knowledge the first to score internal inter-group geometry against
        real survey ground truth: RDM-vs-RDM comparison across layers,
        attention heads, FFN outputs, and prompt variants, with
        winner's-curse-corrected statistics (max-statistic permutation,
        held-out and split-half validation)---and identifies \starhead{}
        as a general demographic-geometry head, significant across all
        six attribute types as a \emph{fixed} location, with fidelity
        heterogeneous by type: political/socio-economic structure is
        encoded at $\rrho$ up to $0.63$, while racial$\times$religious
        structure is weak and prompt-fragile everywhere.
  \item A dissociation of causal use from representational fidelity.
        Replacing the entire prompt identity moves the model's predicted
        distribution by under $2\%$ of its prediction error in aggregate
        (Sec.~\ref{sec:gap}), and a causal sweep over all 32 layers in all
        six types---with cluster-robust, selection-corrected
        inference---shows that the clearest causal pathway sits in a
        low-fidelity type while the most faithful type shows no
        correction-surviving single-layer pathway (Sec.~\ref{sec:sweep}).
  \item A direct measurement of what a faithful read-out buys, with a
        practical reframing. A probe on \starhead{} alone (128 dimensions)
        is $21$--$31\%$ closer to survey truth than the model's own
        answers, but recovers almost no per-question group ordering: the
        two measure different resolutions, faithful \emph{on average} but
        not sharp enough for any single question (Sec.~\ref{sec:probe}).
        If faithful structure already exists internally, surface
        fine-tuning \citep{subpop2025} may re-teach what the model already
        encodes---a candidate mechanism for its known transfer
        failure---and reading the location directly is the cheaper
        alternative, though our probe results show it is still not a
        substitute for survey data.
\end{itemize}

\section{Related Work}
\label{sec:related}

\paragraph{LLMs as survey simulators (the ground-truth side).}
``Silicon sampling'' was introduced by \citet{argyle2023} and has since
been contested: response instability and survey-format sensitivity
\citep{dominguez2024, rottger2024}, systematic divergence from human
samples \citep{bisbee2024}, and bounded persona effects
\citep{hucollier2024}. SubPOP \citep{subpop2025} fine-tunes LLMs to
predict per-group response distributions with a KL objective, with strong
in-distribution but degraded out-of-survey accuracy; rectification methods
\citep{rectification2025} combine limited human data with synthetic
responses. Cultural-alignment work compares model
\emph{outputs} to cross-national survey structure---GlobalOpinionQA
against World Values Survey and Pew Global Attitudes items
\citep{globalopinions2023}, and ChatGPT responses against Hofstede's
dimensions \citep{culturalalign2023}. All treat the model as a black box.

\paragraph{Reading opinions from inside (the closest neighbor).}
\citet{llmopinions2026} show linear probes on internal activations recover
per-group opinion distributions \emph{better than the model's own
generated outputs}, extract activations down to attention heads, and
steer outputs via SAE patching. All their measurement is
\emph{first-order}: each group's read-out is scored against that group's
own distribution. Whether the \emph{inter-group geometry} at any location
mirrors real inter-group opinion structure---the property generalization
depends on---is not among their measurements.\footnote{Their four reported
methods---linear probing, SAE feature analysis, SAE-based steering, and
fine-tuning---are all first-order, per-group comparisons (ICLR 2026
camera-ready, OpenReview \texttt{kHVzEjThKE}; repository
\texttt{schang-lab/llm-opinions}); no inter-group RSA or distance-matrix
analysis is described.} Our work is
the second-order complement: probe accuracy
is a phone book, accurate per entry; we ask whether the entries are
\emph{arranged as a map}. Two works come closest:
\citet{personamanifold2026}
correlate persona-conditioned final-layer separation with human subgroup
disagreement, and \citet{beyondmarginals2026}
test whether aligned models reproduce human item--item correlation beyond
marginals---both compare behavioral or last-layer structure, not internal
geometry against a human inter-group distance matrix. Layer-wise RSA of
LLM geometry against human similarity structure exists for
\emph{perceptual} domains, peaking
mid-stack and fading with depth \citep{perceptualgeom2026}---a parallel to
our demographic-identity finding.

\paragraph{Localizing social attributes (the interpretability side).}
\citet{kimevans2025} localize political ideology to attention heads and
validate against legislator vote-based scores (DW-NOMINATE)---our closest
methodological relative (one attribute, elite text); a follow-up locates a
single partisan axis mid-stack and steers along it
\citep{partisandir2026}.
\citet{personalocal2025,sociodemo2025,personapatch2025} localize persona
and socio-demographic representations---via linear probing across layers
and activation patching---without survey ground truth.
Per-layer RSA applies to LLMs in other content
domains \citep{assoc2026}. ``Diversity collapse'' \citep{divcollapse2026}
documents social-identity representations becoming less separable with
depth---consistent with our anisotropy findings---but not fidelity to real
opinion structure. Construct-validity critiques of
persona prompting \citep{constructvalidity2026} motivate our multi-cue
design. \citet{personaprefs2026} find preference machinery largely shared
across personas---probes and steering vectors transfer between opposed
personas---converging with our dissociation between what a location
encodes and what the model answers with.

\paragraph{The behavioral Sim2Real gap (the motivation side).}
The Sap-group program documents from outside that LLM simulators are
homogeneous and overly cooperative: 451-human comparisons show the best of
31 simulators reaches only USI 76.0 vs.\ a 92.7 human baseline
\citep{sim2real2026}, and OdysSim \citep{odyssim2026} attributes the gap
to helpfulness-driven post-training and rebuilds a behavioral foundation
model to close it---noting that better \emph{average}
human simulation does not imply faithful \emph{subpopulation} simulation.
Persona effects on moral stances are strongest for political ideology and
personality \citep{socratic2025}. Foundational annotation work
\citep{socialchem2020,annotators2022} established that human judgments are
distributions conditioned on who answers---our evaluation's premise.
None of this line opens the model.

\section{Method}
\label{sec:setup}

\subsection{Survey-side ground truth}
We use OpinionQA-style per-group response distributions derived from Pew
American Trends Panel waves \citep{opinionqa2023, subpop2025} (15 waves;
inventory and preprocessing in Appendix~\ref{app:data}). A \emph{group} is an intersectional cell: a pair
of demographic attribute values (e.g., \textit{race=Asian $\times$
religion=Hindu}), yielding $169$ cells across six attribute types
(AGE$\times$POLPARTY, EDUCATION$\times$INCOME, RACE$\times$RELIG,
RACE$\times$POLPARTY, RACE$\times$POLIDEOLOGY, RELIG$\times$POLPARTY).
These six pairs are a purposive, not exhaustive, sample of the
$\binom{12}{2}=66$ possible attribute pairs in this corpus, chosen to
instantiate an intersectional dimension prior work flags as an open
question and to span distinct social/political axes (full justification
in Appendix~\ref{app:sixpairs}). By number of cells clearing the
$n\geq30$ threshold, our six rank 7th to 44th among all 66 pairs, so
selection was not biased toward the best-populated ones
(Appendix~\ref{app:sixpairs}). The pairs were fixed, and the underlying cell
data built, before any fidelity or causal result was computed on them.
For groups $i,j$ the ground-truth distance is the mean Wasserstein
distance between their answer distributions over shared questions
(up to 300 sampled per pair):
\begin{equation}
D^{\mathrm{real}}_{ij} \;=\; \frac{1}{|Q_{ij}|}\sum_{q\in Q_{ij}}
\wdist\!\big(p_i^{(q)},\, p_j^{(q)}\big),
\end{equation}
where $p_i^{(q)}$ is group $i$'s real response distribution on question
$q$'s ordinal scale. $D^{\mathrm{real}}$ is the survey-side RDM.
Appendix~\ref{app:dreal-sens} reports sensitivity of the headline results
to two construction choices: normalising WD by each question's scale
range, and restricting to questions answered by every cell of a type.

\subsection{Model-side representations and prompts}
For each group we build identity prompts and extract, at the final token,
(i) the residual stream after every layer ($33$ locations), (ii) each
attention head's output before the out-projection
($32\times32=1{,}024$ locations; the same read point used by
\citealp{llmopinions2026}), and (iii) each FFN block output ($32$
locations), from Mistral-7B \citep{mistral7b}---our primary model
throughout. The same read-out protocol is reapplied unchanged to a second
model family in Sec.~\ref{sec:crossmodel} (Qwen3-8B-Base and two OdysSim
checkpoints, whose 36-layer stack yields $37+1{,}152+36=1{,}225$
locations by the same construction) and to Mistral-7B-Instruct-v0.2 for
the identity-swap ceiling in Sec.~\ref{sec:ceiling}. The causal
experiments of Sec.~\ref{sec:gap} use activation patching
\citep{meng2022rome, wang2023ioi}, following \citet{heimersheim2024} and
mindful of documented reliability limits \citep{causalprobing2025}. The
model-side RDM at location $\ell$ uses cosine distances between group
representations.

A single prompt template is a single operationalization of ``group
identity'' \citep{constructvalidity2026}. We therefore use four templates
per group (third-person declarative, first-person, structured profile,
and QA format ending in \texttt{Answer:}---the read-out position validated
by \citealp{llmopinions2026}) and additionally a template-averaged
representation (\textsc{Tmean}); the four are reproduced verbatim in
Appendix~\ref{app:templates}. Template disagreement is reported as a
fragility diagnostic.

\subsection{Fidelity: second-order comparison}
Fidelity at location $\ell$ for attribute type $t$ is the Spearman
correlation between the upper triangles of the two RDMs restricted to
type-$t$ cells:
\begin{equation}
\rrho_{\ell,t} \;=\; \mathrm{Spearman}\!\Big(
\big\{D^{\mathrm{real}}_{ij}\big\}_{i<j\in t},\;
\big\{D^{\ell}_{ij}\big\}_{i<j\in t}\Big).
\end{equation}
We report within-type fidelity throughout: pooled correlations mix
between-type scale differences and can both mask weak types and
manufacture spurious gains (Sec.~\ref{sec:act1}).

\subsection{Causal interventions and probes}
\label{sec:setup-causal}
Beyond representational geometry, we test whether outputs causally
depend on it, with four instruments (Appendix~\ref{app:patching};
Appendix~\ref{app:probe}). An identity-swap ceiling measures how much the
predicted opinion distribution moves when the entire demographic
identity is replaced, bounding how identity-responsive the output is at
all. Activation patching at the identity token replaces a donor group's
activations there, propagating the edit forward; our primary
intervention jointly patches all 32 heads of layer 11. A per-layer sweep
repeats this 32-head patch at every layer for all six types; significance
comes from a max-statistic sign-flip permutation test over layers,
computed at the pair level because items cluster into donor--recipient
pairs. Finally, a linear probe fit on \starhead's 128
activations (leave-one-cell-out) asks how much of a group's real
response distribution is linearly recoverable from that location alone.

\subsection{Statistical protection against selection}
\label{sec:setup-selection}
Any ``best location'' claim over $1{,}089$ candidates inflates under
selection (winner's curse). We therefore report three corrections:
(a) \textbf{max-statistic permutation tests} \citep{nichols2002}---the
null distribution of the \emph{maximum} $\rrho$ over all locations under
cell-label permutation; (b) \textbf{held-out-template selection}---locations chosen
on three templates, scored on the fourth; (c) \textbf{split-half
selection}---locations chosen on half the cells, scored on the other half
(200 resamples). Headline numbers are held-out values, not selected
maxima.

Two further disciplines apply to the causal experiments
(Sec.~\ref{sec:gap}). Patching items are \emph{clustered}: each
type contributes 12 donor--recipient pairs $\times$ 20 questions, and
items within a pair share the persona and its idiosyncrasies, so all
causal $p$-values are computed at the \emph{pair} level (sign-flip
permutation over 12 clusters, enumerated exactly;
Appendix~\ref{app:cluster}), with item-level statistics shown only as
descriptive detail. And because the survey-side ruler could itself be
confounded by lexical similarity of the identity words,
Sec.~\ref{sec:act2} reports fidelity after partialling out a
lexical-baseline RDM (Appendix~\ref{app:lexical}).

\section{Results I: Where Demographic Geometry Is Faithful}
\label{sec:act2}

\subsection{The standard residual read-out understates the model}
\label{sec:act1}
The read-out used implicitly across the LLM-survey literature---last-token
residual stream, cosine distance, one prompt template---paints a bleak
picture, but a misleading one. Pooled fidelity is $\rrho\approx0.17$
(peak at layer 17), but this pools unequal types: AGE$\times$POLPARTY
and EDUCATION$\times$INCOME reach $\rrho\approx0.50$ while
RACE$\times$RELIG is $\rrho=0.07$ (n.s.), and cross-type pairs correlate
\emph{negatively}, so distances are only meaningful within a type.
Group embeddings are severely anisotropic
($\lVert\bar{x}\rVert/\overline{\lVert x\rVert}=0.95$): mean-centering
doubles the pooled correlation ($0.185\!\to\!0.301$) yet \emph{decreases}
within-type correlation in all three types measured---a between-type
scale artifact we do not adopt anywhere in this paper
(Appendices~\ref{app:centering} and~\ref{app:anisotropy}). Top-5 neighbor
precision against ground truth is ${\sim}2\times$ chance in all six
types but only $0.35$--$0.48$ absolute, and applications that borrow
statistical strength from neighbors inherit this noise: a
consistency-regularization pilot built on this read-out failed
accordingly (Appendix~\ref{app:lgroup}).

Weakness of \emph{one} read-out does not establish weakness of the
\emph{model}: the residual stream sums all attention heads and FFN
blocks, so signal carried by a few components can be diluted by the
rest. The next subsection reads every component separately.

\subsection{Attention heads dominate the residual stream}
\begin{figure}[t]
\centering
\includegraphics[width=\linewidth]{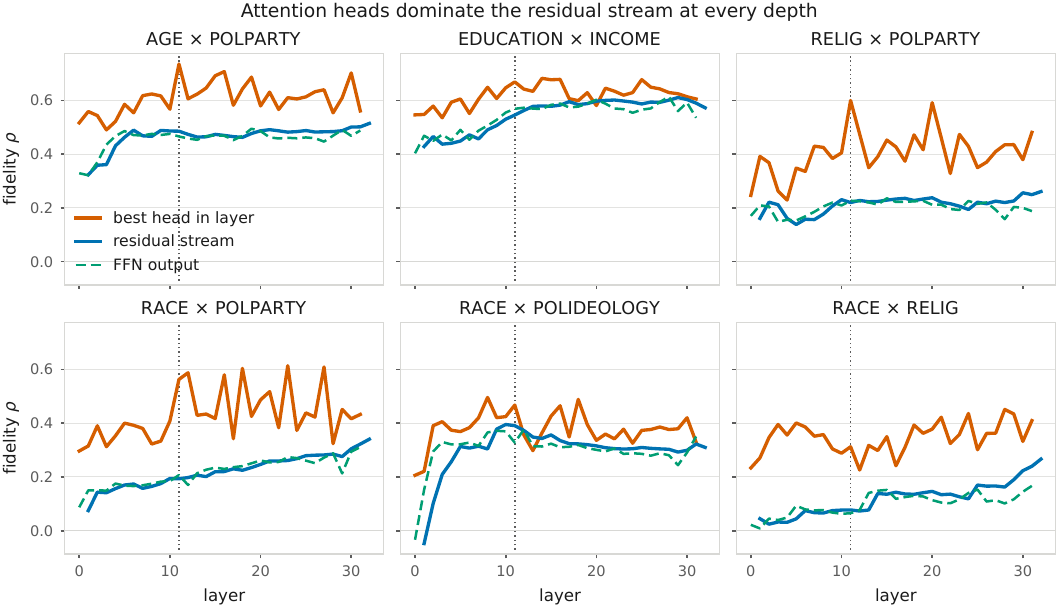}
\caption{Fidelity by depth and component (multi-cue \textsc{Tmean}
read-out): orange = best of 32 heads per layer, blue = residual stream,
dashed green = FFN output. Dotted line marks layer 11. Maxima here are
selected/uncorrected; Table~\ref{tab:map} gives held-out values.}
\label{fig:layercurve}
\end{figure}

The best attention-head read-out exceeds the best residual-stream
read-out in every attribute type, with the largest gains where the
standard read-out was weakest (RELIG$\times$POLPARTY:
$0.26\!\to\!0.60$; RACE$\times$POLPARTY: $0.34\!\to\!0.61$;
Table~\ref{tab:map}); under the symmetric correction of
Sec.~\ref{sec:symmetric} this holds in five of six types, reversing in
none. Multi-cue averaging alone already helps (RACE$\times$RELIG
residual: $0.07\!\to\!0.27$ \textsc{Tmean}), so part of the apparent
weakness was operationalization noise. Figure~\ref{fig:layercurve} shows
this is not one lucky depth: the best head within a layer beats that
layer's residual stream at nearly every depth in all six types (residual
wins at only 3 of $186$ layer--type points), while the FFN output tracks
the residual stream rather than the heads---consistent with dilution at
the summation, not a failure to encode the signal.

Those per-layer curves take the \emph{best} head at each depth and are
selection-contaminated; averaging $\rrho$ over all 32 heads of a layer
instead gives a selection-free view of which depths carry the structure
(Table~\ref{tab:layerblock} in Appendix~\ref{app:quarter-table}). The
first quarter of the stack is uniformly the weakest block, and the peak
is not uniformly mid-stack: it emerges latest for RACE$\times$RELIG,
whose block means rise monotonically into the final quarter---the
weakest type is thus also the slowest to build up.

\begin{table}[t]
\centering
\caption{Fidelity ($\rrho$ vs.\ survey ground truth) by read-out.
\emph{Selected} = maximum over locations; \emph{held-out} =
selection-corrected range from two independent corrections (held-out
template, Appendix~\ref{app:selection}; split-half median), on the
NaN-cleaned cell set.}
\label{tab:map}
\begin{tabular}{lcccc}
\toprule
Type & Residual best & Head best (selected) & Head (held-out) & Best loc.\\
\midrule
AGE$\times$POLPARTY      & $+0.51$ & $+0.74$ & $+0.59$--$0.63$ & L11 H19\\
EDUCATION$\times$INCOME  & $+0.61$ & $+0.68$ & $+0.58$--$0.63$ & L14 H1\\
RELIG$\times$POLPARTY    & $+0.26$ & $+0.60$ & $+0.50$--$0.52$ & \starhead\\
RACE$\times$POLPARTY     & $+0.34$ & $+0.61$ & $+0.41$--$0.52$ & L23 H13\\
RACE$\times$POLIDEOLOGY  & $+0.40$ & $+0.50$ & $+0.30$--$0.41$ & L8 H21\\
RACE$\times$RELIG        & $+0.27$ & $+0.45$ & $+0.21$--$0.32$ & L28 H0\\
\bottomrule
\end{tabular}
\end{table}

\subsection{Controls: selection, geometry, and lexical confounds}
\label{sec:symmetric}
Max-statistic permutation tests (Sec.~\ref{sec:setup-selection}) show
pure noise picking its champion among 1{,}024 heads reaches only
$\rrho\approx0.34$--$0.46$ (95th null percentile), exceeded by the
observed maxima in all six types ($p<0.0005$ in four; $0.0025$,
$0.0155$ in the rest). Held-out-template and split-half estimates agree
(Table~\ref{tab:map}); head \emph{choice} is stable under resampling for
most types (\starhead{} re-selected in 87/200 splits for
RELIG$\times$POLPARTY) but a lottery for RACE$\times$RELIG.

\paragraph{Symmetric correction and a geometry control.} That comparison
is also unequal in dimensionality: 128-dimensional heads might beat the
anisotropic 4{,}096-dimensional stream for geometric, not informational,
reasons. One pipeline with identical selection over three
families---1{,}024 heads, 33 residual layers, and 1{,}024 random
128-dimensional projections of the residual stream (Appendix~\ref{app:randproj})---shows
heads winning in five of six types, with projections tracking the
residual stream rather than the heads, so the advantage is not a
dimensionality artifact; the exception, EDUCATION$\times$INCOME, has a
thin margin, where the standard read-out is already close to the best we
find (Table~\ref{tab:controls}).

\paragraph{A ceiling for fidelity.} A correlation of $0.63$ means
something different under a ceiling of $0.70$ than under $0.95$.
Survey-side split-half reliability is $0.985$--$0.994$; model-side
reliability across template halves at \starhead{} is $0.72$--$0.94$,
giving an attenuation ceiling $\sqrt{r_{\text{survey}}\,r_{\text{model}}}
= 0.84$--$0.96$ per type (observed fidelity reaches $39$--$74\%$ of it,
Table~\ref{tab:controls}): the unexplained residual is real signal the
representation lacks, not an artifact of a noisy ruler.

\begin{table}[t]
\centering\footnotesize
\setlength{\tabcolsep}{4pt}
\caption{Every control in Sec.~\ref{sec:act2}, per attribute type.
$p_{\text{sel}}$: selection-corrected max-statistic $p$. head/resid/proj:
held-out fidelity of the three families (proj = random 128-d
projections). \% ceil: \starhead{} fidelity as a fraction of the
attenuation ceiling. partial $\rrho$: \starhead{} fidelity partialling
out the lexical-baseline RDM.}
\label{tab:controls}
\begin{tabular}{lccccccc}
\toprule
& \multicolumn{1}{c}{selection} & \multicolumn{3}{c}{symmetric + geometry (held-out)}
& \multicolumn{1}{c}{ceiling} & \multicolumn{2}{c}{lexical}\\
\cmidrule(lr){2-2}\cmidrule(lr){3-5}\cmidrule(lr){6-6}\cmidrule(lr){7-8}
Type & $p_{\text{sel}}$ & head & resid & proj & \% ceil & partial $\rrho$ & $p$\\
\midrule
AGE$\times$POLPARTY & $<$.0005 & 0.52 & 0.39 & 0.42 & 74\% & 0.59 & $<$.0005 \\
EDUCATION$\times$INCOME & $<$.0005 & 0.54 & 0.51 & 0.49 & 70\% & 0.35 & $<$.0005 \\
RELIG$\times$POLPARTY & $<$.0005 & 0.44 & 0.17 & 0.18 & 68\% & 0.59 & $<$.0005 \\
RACE$\times$POLPARTY & $<$.0005 & 0.47 & 0.28 & 0.30 & 69\% & 0.58 & $<$.0005 \\
RACE$\times$POLIDEOLOGY & .0025 & 0.38 & 0.26 & 0.29 & 53\% & 0.37 & $<$.0005 \\
RACE$\times$RELIG & .0155 & 0.29 & 0.16 & 0.17 & 39\% & 0.23 & .056 \\
\bottomrule
\end{tabular}
\end{table}

\paragraph{Not an artifact of the identity words themselves.}
\label{sec:lexical}
The prompts differ across cells only in the attribute-value words
(``Democrat'', ``Hindu'', ``30--49'', \dots), whose lexical similarity
correlates with real opinion similarity, so the map could be a mere copy
of word similarity. A baseline RDM built from a small sentence encoder
\citep{minilm2020} (value words alone, and the full identity sentence)
is correlated with survey truth; we report each read-out's
\emph{partial} Spearman correlation after controlling for it, by
cell-label permutation.

The confound is real but does not explain the map: the baseline
alone reaches $\rrho=0.21$--$0.64$ (largest for EDUCATION$\times$INCOME,
whose value words are ordered), yet \starhead's fidelity controlling for
it remains $\rrho=0.35$--$0.59$ ($p<0.0005$) in five of six types, with
the largest drop for EDUCATION$\times$INCOME ($0.67\!\to\!0.35$),
consistent with its word-order structure. The exception is again
RACE$\times$RELIG: partial fidelity there is $0.23$ ($p=0.056$; the
type's best head survives at $0.41$, $p=0.0015$)---the fourth control
this type is weakest under (Table~\ref{tab:controls}). For word-ordered
types, though, some of the raw correlation is lexical.

\subsection{A general demographic-geometry head}
\begin{figure}[t]
\centering
\includegraphics[width=0.66\linewidth]{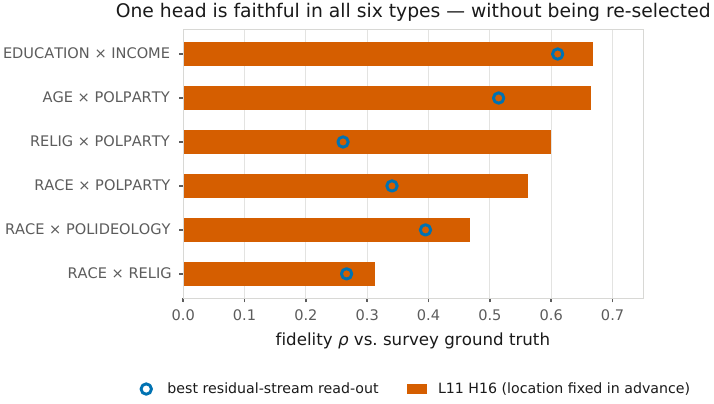}
\caption{Fidelity of the fixed location \starhead{} (bars, never
re-selected per type) against the best residual-stream read-out for
that type (circles); margins $+0.05$ to $+0.34$. Significance ($p<0.001$
each, $\times1024$ Bonferroni) is on the NaN-cleaned cell subset, where
race-containing values shift slightly (RACE$\times$RELIG
$0.31\!\to\!0.33$; RACE$\times$POLPARTY $0.56\!\to\!0.59$; exclusions in
Appendix~\ref{app:data}).}
\label{fig:starhead}
\end{figure}

Fixing \starhead{} in advance (no per-type re-selection) yields
significant fidelity in \emph{all six} types ($\rrho=0.33$--$0.67$;
$p<0.001$ each, surviving a conservative $\times1024$ Bonferroni): a
single head, one of 1{,}024, tracks real opinion structure across age,
education, income, religion, party, ideology, and (weakly) race
(Figure~\ref{fig:starhead}), and, being fixed before
scoring, is structurally immune to the per-type winner's curse of
Sec.~\ref{sec:setup-selection}. One provenance caveat: \starhead{} was
\emph{noticed} for recurring in per-type top-10 lists, so this rules out
per-type selection but not the initial noticing; independent
confirmation comes from the causal test in Sec.~\ref{sec:gap} and the
cross-model replication in Sec.~\ref{sec:crossmodel}, where the
phenomenon recurs at a family-specific address (full head-by-head map in
Appendix~\ref{app:headmap-full}).

\subsection{Heterogeneity across attribute types}
The type hierarchy is preserved at every location tested:
the age- and education-based types are strongly and stably encoded
(template std $0.02$--$0.05$; RELIG$\times$POLPARTY sits between, at
$0.10$), while every race-containing type is weaker \emph{and}
prompt-fragile (template std $0.12$--$0.13$; e.g.,
RACE$\times$RELIG at its best head: $\rrho=0.16$ under one template,
$0.45$ under another). This parallels output-level findings that
political ideology dominates persona effects \citep{socratic2025} and
annotator-attitude findings that ideological variables outpredict raw
demographics \citep{annotators2022}. Demographic attributes therefore
cannot be treated as interchangeable: a persona pipeline validated on
political or socio-economic attributes gives no warrant for race-based
attributes, where this model's internal structure is weaker and unstable
under paraphrase.

\subsection{Cross-model replication}
\label{sec:crossmodel}
The sensitivity analyses above are internal checks; this subsection adds
an external one: replicating the fidelity map on three checkpoints
differing only in human-simulation training \citep{odyssim2026}---Qwen3-8B-Base,
OSim-8B-Mid (midtrained on a 10B-token behavioral corpus), and OSim-8B
(post-RL)---with the same prompts, cells, and pipeline, validated by
reproducing the Mistral map from raw activations.
\phantomsection\label{sec:robustness}

\begin{table}[t]
\centering\footnotesize
\setlength{\tabcolsep}{3pt}
\caption{Cross-model replication of the fidelity map. \emph{head/resid} =
selected best head vs.\ best residual read-out. $\rrho_{\text{ho}}$ =
held-out-template fidelity (max-statistic $p$ in parentheses,
$2{,}000$ permutations). Mistral column: that model's held-out head
value, for reference.}
\label{tab:crossmodel}
\begin{tabular}{lccccccc}
\toprule
& Mistral & \multicolumn{2}{c}{Qwen3-8B-Base}
& \multicolumn{2}{c}{OSim-8B-Mid} & \multicolumn{2}{c}{OSim-8B (post-RL)}\\
\cmidrule(lr){2-2}\cmidrule(lr){3-4}\cmidrule(lr){5-6}\cmidrule(lr){7-8}
Type & $\rrho_{\text{ho}}$ & head/resid & $\rrho_{\text{ho}}$ ($p$)
& head/resid & $\rrho_{\text{ho}}$ ($p$)
& head/resid & $\rrho_{\text{ho}}$ ($p$)\\
\midrule
AGE$\times$POLPARTY & 0.52 & 0.78/0.50 & 0.49\,($<$.0005) & 0.74/0.50 & 0.61\,($<$.0005) & 0.73/0.51 & 0.57\,($<$.0005) \\
EDUCATION$\times$INCOME & 0.54 & 0.70/0.51 & 0.58\,($<$.0005) & 0.68/0.50 & 0.54\,($<$.0005) & 0.68/0.53 & 0.60\,($<$.0005) \\
RELIG$\times$POLPARTY & 0.44 & 0.57/0.37 & 0.56\,($<$.0005) & 0.58/0.39 & 0.56\,($<$.0005) & 0.56/0.38 & 0.55\,($<$.0005) \\
RACE$\times$POLPARTY & 0.47 & 0.60/0.37 & 0.37\,(.001) & 0.57/0.38 & 0.38\,(.002) & 0.63/0.37 & 0.46\,($<$.0005) \\
RACE$\times$POLIDEOLOGY & 0.38 & 0.59/0.35 & 0.38\,($<$.0005) & 0.62/0.36 & 0.38\,($<$.0005) & 0.60/0.35 & 0.40\,($<$.0005) \\
RACE$\times$RELIG & 0.29 & 0.41/0.35 & 0.29\,(.15) & 0.45/0.36 & 0.30\,(.062) & 0.43/0.33 & 0.22\,(.096) \\
\bottomrule
\end{tabular}
\end{table}

Four results: heads dominate the residual stream here too (best head
beats best residual in all six types at all three checkpoints, tracking
Mistral's magnitudes, e.g.\ RELIG$\times$POLPARTY $0.56$ vs.\ $0.44$);
the type hierarchy replicates, race weakness included (five of six types
pass max-statistic correction at every checkpoint, $p\le0.0025$; the one
failure, always RACE$\times$RELIG, was Mistral's weakest too); a general
demographic-geometry head recurs at a family-specific address (L33\,H9,
fixed across types and checkpoints, scores $\rrho=0.35$--$0.58$ in five
types, weakest again on RACE$\times$RELIG); and ten billion tokens of
simulation training barely move the map or the mouth (base$\to$post-RL
held-out fidelity shifts $-0.07$ to $+0.09$ with no systematic
direction, output accuracy changes $\le0.03$, output-RDM fidelity
improves $0.04$)---consistent with the OdysSim authors' own caution that
better average-human simulation does not imply better subpopulation
simulation.

The causal side does not yet replicate: patching (Sec.~\ref{sec:gap})
has only been run on Mistral, so the map--use dissociation remains a
one-model result.

\section{Results II: Whether the Faithful Geometry Is Used}
\label{sec:gap}

\subsection{Outputs are nearly identity-invariant under natural prompting}
\label{sec:ceiling}
Before asking whether any internal location causally drives answers, we
measure the ceiling: how much does the model's predicted opinion
distribution move when the \emph{entire} demographic identity in the
prompt is replaced? Averaged over pairs and questions, the answer is
almost not at all: $\wdist(\hat{p}_A, p^{\mathrm{real}}_B) -
\wdist(\hat{p}_B, p^{\mathrm{real}}_B) \approx 0.004$ against a total
prediction error of ${\sim}0.37$ (${<}2\%$). (One illustrative item, not
an aggregate statistic: an 11-point real gap between 18--29- and
50--64-year-olds moves by only 1.7 points under a full persona
swap---${\sim}15\%$ of the true difference.)

Worse, the small variation that does exist is directionally
uninformative: the prediction for group $A$ is \emph{no closer} to $A$'s
own ground truth than to another group's (median advantage
$\approx0$; sign test ${\approx}$ coin flip in 2 of 3 tested types).
The model produces slightly-different-but-equally-wrong answers per
persona. 

We re-measured this ceiling on the instruction-tuned
Mistral-7B-Instruct-v0.2 under its own chat template: the near-invariance
replicates, with the instruct model's full-identity swap moving
predictions by $3.9\%$ of its total error (Appendix~\ref{app:ceiling-instruct};
Table~\ref{tab:ceiling}). This near-invariance is a property of the
model family, not of the raw-completion interface, though both estimates
assume the letter-probability read-out (Sec.~\ref{sec:limitations}).

The aggregate also hides a large per-type spread that inverts the
fidelity ranking at the two extremes (Appendix~\ref{app:ceiling-spread}).

\subsection{Patching at layer 11: a real but small effect}
\label{sec:weakinstrument}
Patching \starhead{} activations between group prompts (donor$\to$base) at
the \emph{prediction} token, with 1--3 heads at natural scale, moves
nothing (protocol in Appendix~\ref{app:patching}). But the ceiling above
shows this instrument had a dynamic range of under $2\%$ of total
prediction error to begin with---any causal effect this small would be
undetectable regardless of whether it exists.

\label{sec:v2}
We therefore repeat the intervention with an instrument matched in
strength to prior successful steering work \citep{llmopinions2026}: all
32 heads of layer 11 are patched jointly at a single identity token, so
the edit propagates to every later token (protocol in
Appendix~\ref{app:patching}). Across the three types tested here the
result is \textbf{graded}, and the gradation appears to line up with
each type's fidelity from Sec.~\ref{sec:act2}; Sec.~\ref{sec:sweep}
extends the experiment to all six types and shows that this alignment
does \emph{not} generalise.

\begin{table}[t]
\centering
\caption{Causal patching results with the strong instrument. Ceiling =
mean full-prompt-swap shift (Sec.~\ref{sec:ceiling}, this instrument's
question set). Shift = mean directional shift toward the target group's
true distribution; \% of ceiling in parentheses where meaningful.
$p_{\text{item}}$ = Wilcoxon vs.\ random-head control over 240 items
(descriptive; items are clustered); $p_{\text{pair}}$ = exact sign-flip
over the 12 pairs (primary).}
\label{tab:patchv2}
\resizebox{\linewidth}{!}{%
\begin{tabular}{lccccc}
\toprule
Type (split-half fidelity) & Ceiling & 32 heads @ L11 & Single \starhead{} & $p_{\text{item}}$ (32h) & $p_{\text{pair}}$ (32h) \\
\midrule
AGE$\times$POLPARTY ($0.63$)   & $+0.0165$ & $+0.0067$ (41\%) & $-0.0001$ (n.s.) & $0.0027$ & $0.019$ \\
RELIG$\times$POLPARTY ($0.52$) & $+0.0020$ & $+0.0034$ (170\%, n.s.) & $-0.0004$ (n.s.) & $0.16$ & $0.15$ \\
RACE$\times$RELIG ($0.21$)     & $+0.0026$ & $+0.0017$ (n.s.) & $\mathbf{-0.0005}$ (wrong dir., $p_{\text{pair}}{=}1/4096$) & $0.31$ & $0.29$ \\
\bottomrule
\end{tabular}%
}
\end{table}

For AGE$\times$POLPARTY, jointly patching all 32 heads at layer 11 shifts
predictions toward the target group's true distribution: $59\%$ of
trials move in the correct direction (vs.\ $32\%$ under an empirical
random-head control), recovering $41\%$ of the achievable ceiling. The
primary, pair-level test is marginal ($p_{\text{pair}}=0.019$; adding
layer 18: $0.016$) and does not survive a Holm correction over the
family of 24 tests behind Table~\ref{tab:patchv2}. Identity is thus only
\emph{plausibly} load-bearing here; Sec.~\ref{sec:sweep} localises
further.

For RACE$\times$RELIG, the 32-head patch is not significant
($p_{\text{pair}}=0.29$), while patching \starhead{} \emph{alone} shifts
predictions in the \emph{wrong} direction with the most consistent effect
in the table (worse than the random-head control in all 12 pairs;
$p_{\text{pair}}=1/4096$). A donor-control experiment shows this
backfire requires a coherent identity donor rather than merely a large
disruption to the representation (Appendix~\ref{app:amplification}).

Fidelity and single-unit causal sufficiency are dissociable:
\starhead{} alone never produces a correct, significant shift in any
type, and a correct effect requires the joint action of all 32
heads---faithful as a representation, but insufficient (or actively
misleading) as a sole intervention target. Scaling the patch further
backfires rather than strengthens it (Appendix~\ref{app:amplification}).

\subsection{The causal locus does not track fidelity}
\label{sec:sweep}
Section~\ref{sec:v2} localises an effect at layer 11 for one type and
finds none for another---too few types to tell whether causal use tracks
fidelity at all. We therefore replace the hypothesis-driven probe with
an exhaustive one, the same 32-head identity-token intervention applied
at \emph{every} layer, for all six attribute types (procedure and
selection corrections in Appendix~\ref{app:sweep-procedure}).

\begin{figure}[t]
\centering
\includegraphics[width=\linewidth]{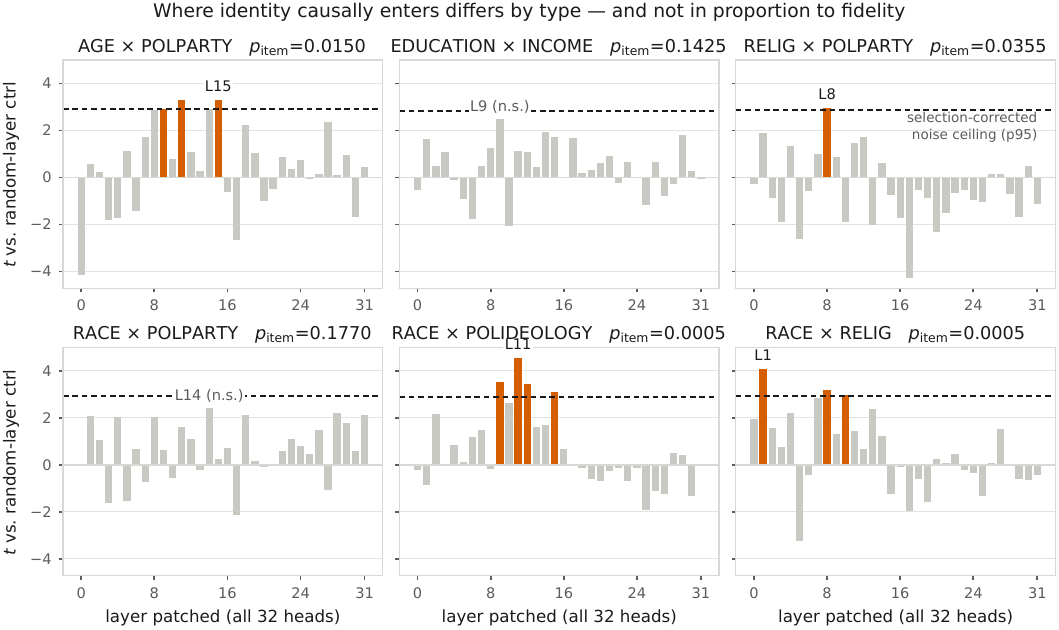}
\caption{Causal effect of patching all 32 heads of each layer in turn, at
the identity token ($n{=}240$ pair--question items per type).
\emph{Descriptive} view: bars are the item-level paired $t$ of the
directional shift toward the target group's true distribution against a
random-layer control; the dashed line is the item-level max-statistic
noise ceiling. Inference in the text and Table~\ref{tab:sweep} is
cluster-robust (sign flips at the level of the 12 pairs).}
\label{fig:sweep}
\end{figure}

\begin{table}[t]
\centering
\caption{Per-layer patching sweep, all six types. Fidelity = split-half
median (one estimator used consistently; see text for sensitivity).
Item-level columns are descriptive ($t$ over 240 items; max-statistic $p$
over 32 layers); the primary inference is the pair-level max-statistic
$p$ (sign flips over 12 pairs, exact) and the pair-level $p$ at the
a-priori locus L11 (no layer selection). The item-level and pair-level
winning layers can differ, since averaging within pairs reweights the
evidence. Output ceiling = mean shift from replacing the entire
identity.}
\label{tab:sweep}
\resizebox{\linewidth}{!}{%
\begin{tabular}{lcccccccc}
\toprule
& & \multicolumn{3}{c}{item-level (descriptive)} & \multicolumn{3}{c}{pair-level (primary)} & \\
\cmidrule(lr){3-5}\cmidrule(lr){6-8}
Type & Fidelity & Win layer & $t$ & $p_{\max}$ & $t_{\text{win}}$ & $p_{\max}$ & $t$@L11 $\rightarrow$ $p$ & Out.\ ceiling \\
\midrule
RACE$\times$POLIDEOLOGY  & $0.30$ & L11 & $4.55$ & $0.0005$ & $3.22$ & $0.066$ & $\mathbf{3.22 \rightarrow 0.0020}$ & $0.0176$ \\
RACE$\times$RELIG        & $0.21$ & L1  & $4.10$ & $0.0005$ & $2.94$ (L8) & $0.122$ & $0.68 \rightarrow 0.27$ & $0.0026$ \\
AGE$\times$POLPARTY      & $0.63$ & L15 & $3.31$ & $0.0150$ & $2.37$ (L14) & $0.266$ & $1.93 \rightarrow 0.022$ & $0.0165$ \\
RELIG$\times$POLPARTY    & $0.52$ & L8  & $2.94$ & $0.0355$ & $2.58$ (L8) & $0.210$ & $1.11 \rightarrow 0.15$ & $0.0020$ \\
EDUCATION$\times$INCOME  & $0.58$ & L9  & $2.47$ & $0.1425$ & $1.57$ (L15) & $0.822$ & $0.75 \rightarrow 0.23$ & $0.0194$ \\
RACE$\times$POLPARTY     & $0.52$ & L14 & $2.40$ & $0.1770$ & $2.00$ (L4) & $0.524$ & $1.28 \rightarrow 0.11$ & $0.0032$ \\
\bottomrule
\end{tabular}%
}
\end{table}

Figure~\ref{fig:sweep} shows the descriptive item-level picture; the
pair-level statistics behind Table~\ref{tab:sweep} are conservative (12
clusters per type) and prune it sharply. No type survives the pair-level
max-statistic correction (best: RACE$\times$POLIDEOLOGY, $p=0.066$). The
a-priori test at L11 (no layer selection; fixed by Sec.~\ref{sec:act2})
is cleaner: RACE$\times$POLIDEOLOGY reaches $t=3.22$, exact $p=0.0020$,
the only entry that also survives a Bonferroni correction over the 12
fixed-location tests below (Table~\ref{tab:fixedlayer}); AGE$\times$POLPARTY
at L11 ($p=0.022$) is suggestive but not conclusive. The item-level loci
otherwise cluster mid-stack and at layer~1 (selection caveat in
Appendix~\ref{app:sweep-procedure}).

\begin{table}[t]
\centering
\caption{Fixed-location causal test, pair-level (exact sign-flip over 12
pairs, one-sided). L11 was fixed by Sec.~\ref{sec:act2} before any
sweep; L1 is pre-registered only for the upper block (see text).
Bonferroni for the 12 tests is $0.0042$; the single bold entry survives
it.}
\label{tab:fixedlayer}
\begin{tabular}{lcccc}
\toprule
Type & $t$ @ L11 & $p_{\text{pair}}$ & $t$ @ L1 & $p_{\text{pair}}$ \\
\midrule
\multicolumn{5}{l}{\emph{types not in the sweep that suggested L1:}} \\
RACE$\times$POLIDEOLOGY  & $\mathbf{3.22}$ & $\mathbf{0.0020}$ & $-0.74$ & $0.77$ \\
RACE$\times$POLPARTY     & $1.28$ & $0.114$ & $1.97$ & $0.033$ \\
EDUCATION$\times$INCOME  & $0.75$ & $0.234$ & $1.24$ & $0.123$ \\
\midrule
\multicolumn{5}{l}{\emph{types that were in that sweep (L1 column not selection-free):}} \\
AGE$\times$POLPARTY      & $1.93$ & $0.022$ & $0.34$ & $0.386$ \\
RELIG$\times$POLPARTY    & $1.11$ & $0.150$ & $1.46$ & $0.089$ \\
RACE$\times$RELIG        & $0.68$ & $0.267$ & $2.46$ & $0.016$ \\
\bottomrule
\end{tabular}
\end{table}

\begin{figure}[t]
\centering
\includegraphics[width=\linewidth]{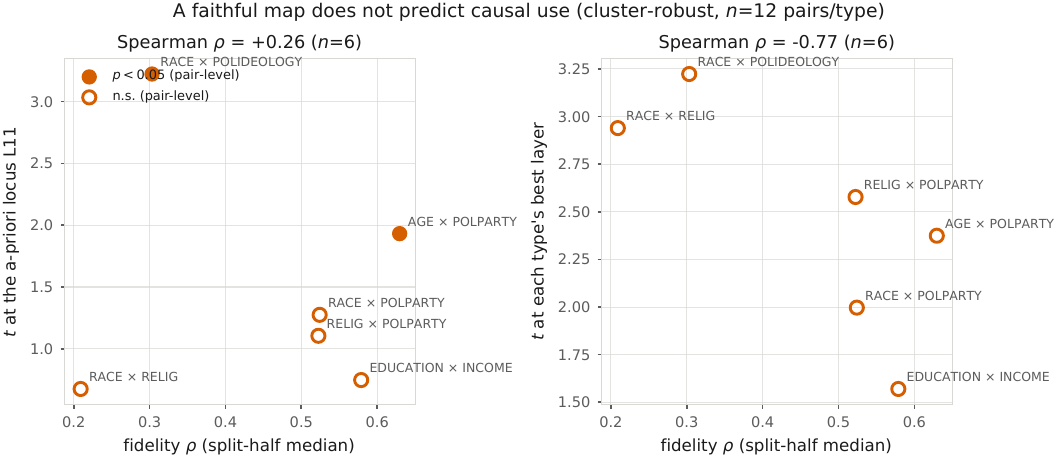}
\caption{Representational fidelity (split-half median) against
cluster-robust causal strength (pair-level $t$), one point per attribute
type. Left: at the a-priori locus L11 (filled = exact pair-level
$p<0.05$). Right: at each type's own best layer (filled = survives the
pair-level max-statistic correction; none does).}
\label{fig:dissoc}
\end{figure}

The natural hypothesis after Sec.~\ref{sec:v2}---that causal strength
should track fidelity---does not survive the extension to six types: the
rank correlation between fidelity and pair-level causal strength is
nowhere significantly positive under any estimator we have, including
the lexical-partial fidelity of Sec.~\ref{sec:lexical} ($+0.26$ to
$-0.77$ depending on ruler and layer, all $p \ge 0.07$; weakly powered at
$n{=}6$, Figure~\ref{fig:dissoc}). What carries the weight is the two
extreme types, tested \emph{directly} rather than by comparing
significance verdicts \citep{gelman2006}: the top-fidelity
types---whichever estimator names them---show no correction-surviving
single-layer locus (EDUCATION$\times$INCOME has the largest output
ceiling and largest raw shift of any type, yet $p=0.82$ at its best
layer; AGE$\times$POLPARTY, the split-half leader, reaches only a
suggestive $p=0.022$ at L11 that fails every correction here), while
RACE$\times$POLIDEOLOGY---among the weakest maps under every
estimator---is the single result that survives everything ($t=3.22$ at
L11, exact $p=0.0020$, surviving Bonferroni, recovering $42\%$ of its
ceiling). Representational fidelity is therefore not what decides
whether identity is causally used: the clearest causal pathway sits in
the type with the poorest map, and the graded pattern of Sec.~\ref{sec:v2}
does not generalise to all six types.

Sec.~\ref{sec:act2}'s fidelity is scored on an identity-only prompt,
whereas the causal and probe experiments are scored in a full QA
context, so a tempting alternative reading is that the geometry simply
does not survive into the answering context, not that the model fails
to consult its map. Measured directly in that context---on the same
activations collected for the probe (Sec.~\ref{sec:probe})---the map
does weaken in five of six types,
but the dissociation itself persists under the QA-context ruler at the
same a-priori locus ($\rrho=+0.03$, $p=0.96$), so ``we measured the
wrong map'' does not explain the dissociation away
(Appendix~\ref{app:qa-context-map}).

\subsection{Reading the map directly: probe versus output}
\label{sec:probe}
The causal experiments ask whether the model \emph{uses} what it
encodes; the complementary question is how much there is to use if we
read the faithful location directly rather than listen to the model's
mouth. We fit a ridge probe, evaluated leave-one-cell-out, at three
locations---full layer 11, the single head \starhead{}, and layer
1---against two references: the model's own answer, and a group-blind
baseline built from other cells' real answers on the same question
(full protocol in Appendices~\ref{app:probe-procedure}, \ref{app:probe}).

\begin{figure}[t]
\centering
\includegraphics[width=\linewidth]{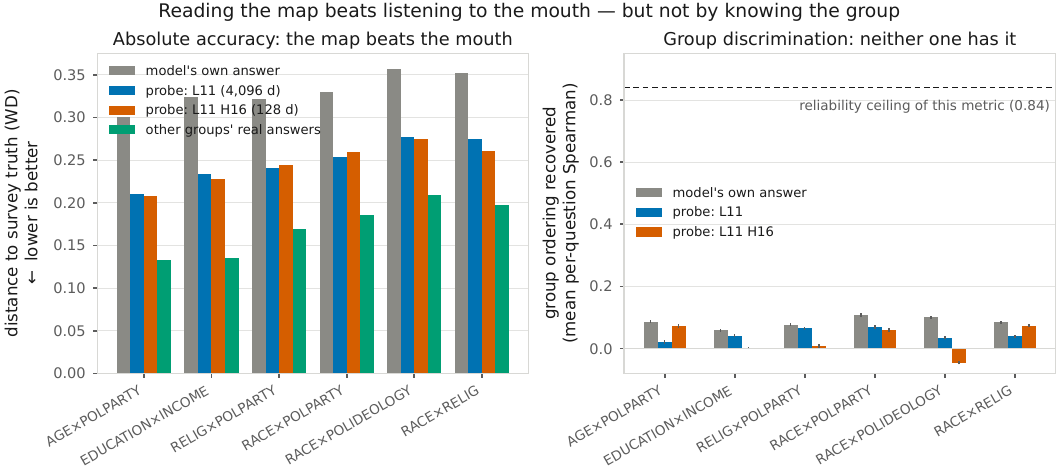}
\caption{Left: distance to survey truth (Wasserstein, lower is better) for
the model's own answer, probes read from the faithful location, and a
group-blind baseline built from other cells' real answers. Right: how much
of the true per-question group \emph{ordering} each read-out recovers
(mean per-question Spearman across cells, half-split evaluation; error
bars are standard errors), against the reliability ceiling of the metric
given survey sampling noise.}
\label{fig:probe}
\end{figure}

In all six types the probe is substantially closer to survey truth than
the model's own answer ($22$--$30\%$ closer overall, better on
$65$--$68\%$ of individual items), not merely by better calibration:
fitting a temperature parameter to the mouth barely helps
($0.300\to0.298$ for AGE$\times$POLPARTY), reproducing against real
survey ground truth the claim that internal states are more informative
than generated answers \citep{llmopinions2026}.

But the gain is question-level, not group-level: a group-blind baseline
that ignores who is being asked and predicts the mean real distribution
of the other cells on the same question beats every read-out we have, by
a wide margin ($0.133$ vs.\ $0.211$ for AGE$\times$POLPARTY; $0.135$
vs.\ $0.234$ for EDUCATION$\times$INCOME). Ranking cells within each
question by predicted opinion and correlating with the true ranking
(half-split, to avoid a leave-one-out artifact) gives a flat null for
everything: the full-layer L11 probe recovers only $+0.02$ to $+0.07$,
the single head \starhead{} is noisier still ($-0.05$ to $+0.07$), and
the model's own answers recover $+0.06$ to $+0.11$---at least as much as
either probe---against a reliability ceiling of $+0.81$ to $+0.86$.

These two results measure different resolutions and are consistent:
Sec.~\ref{sec:act2} scores group geometry aggregated over hundreds of
questions, where the arrangement mirrors survey structure ($\rrho$ up to
$0.63$ held-out); this section scores per-item discrimination, where the
signal is near zero. The model holds a group map that is faithful on
average, but not sharp enough to say who answers what on any particular
question.

If the goal is a per-group opinion distribution, reading $128$
dimensions of one attention head is markedly better than asking the
model, but still worse than using other groups' real answers to the
same question (practical implications in Sec.~\ref{sec:discussion}).

\section{Discussion}
\label{sec:discussion}

\paragraph{Three properties, three literatures.}
``Can LLMs simulate populations?'' conflates three claims. Prior work
established property (i), \emph{readable} \citep{llmopinions2026}, and
documented the behavioral symptom from outside
(\citealp{sim2real2026,odyssim2026}). We establish property (ii),
\emph{faithfully arranged}---at specific locations, unevenly across
attribute types---and show that property (iii), \emph{causally used}, is
not a function of it. Under cluster-robust inference, causal use is not
absent---one type shows a clear, Bonferroni-surviving pathway at L11, and
two more are suggestive---but it does not follow the map
(Sec.~\ref{sec:sweep}). Natural-prompt simulation therefore does not fail
simply because the map is weak; a good map is no guarantee the model
consults it.

\paragraph{A mechanistic candidate for the Sim2Real gap---and a warning
about explaining it by representation quality.}
Behavioral homogeneity of LLM simulators is well measured from outside
\citep{sim2real2026}. Our results offer a consistent inside account:
outputs vary little across personas overall
(Sec.~\ref{sec:ceiling}), and forcing identity through the network moves
them only slightly under a deliberately strong instrument. What our
results do \emph{not} support is that simulation fails because the
internal representation is poor: homogeneity looks like at least two
mechanisms---insufficient causal throughput where the map is good, and no
good map to throughput where it is not. One qualification: the behavioral
literature evaluates post-trained assistants, whereas we measure a base
model; since post-training is itself a plausible cause of
homogenization, our results establish only that a representational
component of homogeneity predates post-training, rather than explaining
any particular assistant's behavior.

\paragraph{Why surface fine-tuning may not transfer.}
SubPOP-style fine-tuning \citep{subpop2025} improves in-distribution
accuracy but degrades on unseen surveys. If faithful structure already
exists internally, output-surface fine-tuning may re-teach what the model
already encodes---a survey-specific surface mapping instead of a
connection to the existing map. This is a testable hypothesis, not a
demonstrated mechanism; the direct test (does reading from the faithful
location transfer where surface fine-tuning does not?) is future work.

\paragraph{Practical guidance now.}
For population simulation: check second-order fidelity before relating
groups, treat demographic attributes as heterogeneous
(Sec.~\ref{sec:act2}), and measure identity conditioning rather than
assume it. \textbf{Do not steer through a single apparently-faithful unit
without a corruption control}: our causal analysis's most consistent
effect is a \emph{harmful} one from doing exactly this
(RACE$\times$RELIG, \starhead{} alone, wrong direction,
$p_{\text{pair}}=1/4096$; Sec.~\ref{sec:v2}). Read a per-group
distribution rather than ask for it: a linear probe on \starhead{} alone
lands $21$--$31\%$ closer to survey truth than the model's own answers
(Sec.~\ref{sec:probe}), yet a group-blind baseline still wins on
per-question ordering.

\section{Conclusion}
\label{sec:conclusion}

We asked whether a model that can be asked about a demographic group also
faithfully represents, and actually uses, that group's opinion structure.
The answer dissociates. Beyond established readability
\citep{llmopinions2026}, the group geometry is \emph{faithfully arranged}
at specific locations---selection-corrected $\rrho$ up to $0.63$---but
only weakly \emph{causally used}: a full identity swap moves predictions
by under $2\%$ of their error, and a linear probe on the faithful
location lands $21$--$31\%$ closer to survey truth than the model's own
answers, yet recovers almost no per-question ordering of groups. Readable,
faithfully arranged, and causally used are therefore separable properties
of the same model, and treating them as one claim is why ``can LLMs
simulate populations?'' has stayed unresolved: a model can encode a
faithful map of a population and still not consult it when asked. Settling
how general this dissociation is will need cross-model causal replication
and cross-institution ground truth beyond a single survey house
(Sec.~\ref{sec:limitations}).

\section{Limitations}
\label{sec:limitations}

\textbf{Scope of replication.} The representational findings replicate on
three checkpoints of a second model family (Sec.~\ref{sec:crossmodel}),
but the patching experiments---and the map-use dissociation with
them---have so far only run on Mistral-7B; output-side near-invariance
does replicate on an instruction-tuned Mistral (Sec.~\ref{sec:ceiling}),
so it is not a base-model artifact. Every patch is single-layer, so
redundant encoding across layers remains an alternative explanation for a
null result only partially ruled out (Sec.~\ref{sec:sweep}, caveat v).
\starhead{} itself was first noticed in per-type top-10 lists before
being fixed for the all-type test; Sec.~\ref{sec:crossmodel} supplies a
second-model check (L33\,H9, selection-free on five types), but a
preregistered replication is still outstanding.

\textbf{Measurement interface and contexts.} Output distributions are
read from option-letter logits after \texttt{Answer:}---the standard
silicon-sampling interface, but known to diverge from free-text answers
\citep{myanswerisc2024}. Our output-side claims (Sec.~\ref{sec:gap})
therefore scope to this interface: identity barely reaches the
letter-probability read-out, not necessarily free-text generation, which
we did not measure; the representation-side claims
(Sec.~\ref{sec:act2}) do not depend on the read-out at all. Relatedly,
the two sides are measured in different input contexts (identity-only
vs.\ QA), and the map is measurably weaker in the QA context
(Sec.~\ref{sec:sweep})---the fidelity--causality correlation is unchanged
there, but our headline fidelity numbers describe an identity-only prompt
and should be read as such.

\textbf{Ground truth and sample.} All distances are measured against one
US-centric institution, Pew ATP; cross-institution transfer (e.g.,
scoring the map against GSS-derived inter-group distances) and
cross-country replication (e.g., WVS, with a known language confound) are
future work. The dissociation claim rests on six attribute types---a
purposive sample of the 66 possible pairs among our twelve attributes
(Sec.~\ref{sec:setup}, Appendix~\ref{app:data})---enough for two
unambiguous counterexamples but not to generalize to intersections
outside this set, and some cells have thin survey samples, with NaN
pairs excluded (Appendix~\ref{app:data}). With cluster-robust inference
the causal experiments have twelve effective units per type: enough for
one clear result and direct between-type contrasts, not to certify a
sweep winner against 31 rivals or estimate the fidelity--causality
relationship precisely ($n{=}6$ types); scaling pairs, not questions, is
the binding constraint for follow-up work.

\textbf{Correlational map, one lexical encoder.} RSA fidelity is
correlational; the patching program (Sec.~\ref{sec:v2},
Sec.~\ref{sec:sweep}) addresses use, not the map's origin. The
partial-fidelity control (Sec.~\ref{sec:lexical}) uses one small sentence
encoder as the lexical baseline; static word vectors cannot represent
several value strings (e.g.\ age ranges), so a stronger lexical model
would only make the control stricter.

\section*{Ethics Statement}
Simulating the opinions of demographic groups is dual-use: it supports
legitimate survey-methodology research---pilot design, non-response
modelling, coverage checks---but the same capability can manufacture
apparently representative public opinion, or reinforce stereotypes by
presenting a model's guess about a group as that group's view.

Our findings cut against the permissive reading: current LLM outputs are
\emph{not} substitutes for human respondents. Under natural prompting the
model's answers move by less than 2\% of their error when the entire
demographic identity is replaced, and the small movement that exists is
not reliably toward the target group's truth. The failure is also
unequal---race-based attributes are the weakest and most prompt-fragile
everywhere measured, so simulated outputs for racial groups are least
trustworthy where misuse would be most harmful. We therefore caution
against policy-adjacent uses of simulated group opinions.

Interpretability results of this kind carry their own risk: a faithful
location is a natural target for steering, and Sec.~\ref{sec:v2} shows
naive steering through such a location can push predictions \emph{away}
from the truth in low-fidelity types---read the map before trusting it as
a lever.

All data are public, aggregated Pew ATP response distributions with no
personally identifiable information; no human subjects were recruited for
this work.

\section*{Reproducibility Statement}
All experiments use the public checkpoint \texttt{mistralai/Mistral-7B-v0.1}
(fp16), prompts and preprocessing fully specified in
Appendix~\ref{app:data}--\ref{app:templates}, and fixed seeds (42
throughout; permutation analyses additionally state their counts).
Ground truth derives from public, aggregated Pew American Trends Panel
response distributions in the OpinionQA/SubPOP format
\citep{opinionqa2023, subpop2025}; no individual-level data are
redistributed. GPU work ran on freely available 2$\times$T4 instances;
all statistics and figures regenerate from released CSVs via included
scripts (analysis code and a figure generator accompany the repository,
which we will release publicly upon publication).

\bibliographystyle{plainnat}
\bibliography{references}

\appendix
\section{Data and Prompts}

\subsection{Data preprocessing and cell inventory}
\label{app:data}
We use the intersectional OpinionQA table derived from Pew American Trends
Panel microdata \citep{opinionqa2023, subpop2025}: $177{,}217$
group--question rows covering $1{,}494$ questions from $15$ ATP waves
(26, 27, 29, 32, 34, 36, 41, 42, 43, 45, 49, 50, 54, 82, 92). Each row
carries a group's response distribution over that question's ordinal
scale; the option list is truncated to the ordinal length, dropping the
trailing \textit{Refused} category.

A \emph{cell} is one intersection of two attribute values within one
attribute type. The six types contribute $169$ cells in total
(Table~\ref{tab:cellinventory}):

\begin{table}[t]
\centering
\begin{tabular}{lcc}
\toprule
Attribute type & Cells & Pairs \\
\midrule
AGE$\times$POLPARTY      & 18 & 153 \\
EDUCATION$\times$INCOME  & 33 & 528 \\
RACE$\times$POLIDEOLOGY  & 30 & 435 \\
RACE$\times$RELIG        & 30 & 435 \\
RELIG$\times$POLPARTY    & 30 & 435 \\
RACE$\times$POLPARTY     & 28 & 378 \\
\bottomrule
\end{tabular}
\caption{Cell and same-type pair counts per attribute type. \emph{Pairs}
= $\binom{\text{cells}}{2}$, the number of RSA observations available for
that type.}
\label{tab:cellinventory}
\end{table}

The survey-side distance $D^{\mathrm{real}}_{ij}$ averages the
Wasserstein distance over questions answered by \emph{both} cells, capped
at $300$ questions sampled per pair (seed 42); questions whose ordinal
scale differs in length between the two cells are skipped. Pairs with no
shared questions stay \texttt{NaN} and are excluded pairwise from RSA; the
selection-correction analyses (Sec.~\ref{sec:setup-selection}) additionally
drop whole cells that produce \texttt{NaN} pairs, which removes 5--6 cells
in the race-containing types and shifts their $\rrho$ values by
$\lesssim0.04$. The patching experiments further restrict to questions
with at most six options (so the answer letters fit \texttt{A}--\texttt{F})
and to pairs sharing at least 20 such questions. Median cell sample size
is $n_{\text{unweighted}}=132$ (minimum 30).

\subsection{Why these six attribute pairs}
\label{app:sixpairs}
Sec.~\ref{sec:setup} explains in one sentence why these six pairs were
chosen; the full rationale and a data-richness check follow.

Two (RACE$\times$RELIG, RACE$\times$POLPARTY) directly instantiate a
dimension the closest prior work names as an open question: how to
``ensure that LLMs faithfully capture opinions along other dimensions
that are not explored in this work, such as intersections of demographic
identities'' \citep[p.~9]{subpop2025}---in our cell set, e.g.,
\textit{White Mormon}, \textit{Asian Republican}. The remaining four were
chosen to span distinct social/political axes rather than for data
richness.

This choice was not, however, biased toward the best-populated pairs, as
the following ranking check shows. The twelve demographic attributes in
this corpus admit $\binom{12}{2}=66$ possible attribute pairs. Applying
the same per-wave $n\geq30$ rule used above to all 66---counting, for
each pair, the number of distinct attribute-value combinations that ever
clear the threshold in any of the 15 waves---ranks our six pairs 7th to
44th of 66 by cell count (Table~\ref{tab:pair-rank}). The richest pair by
this measure (INCOME$\times$RELIG, 39 cells) and the poorest
(CITIZEN$\times$SEX, 5 cells) were never tested. Selection was therefore
not biased toward the best-populated pairs; the full ranking script and
output accompany the repository.

\begin{table}[t]
\centering
\begin{tabular}{lcc}
\toprule
Attribute type (used) & Cells & Rank of 66 (tied block) \\
\midrule
EDUCATION$\times$INCOME  & 33 & 7th (7--8) \\
RACE$\times$POLIDEOLOGY  & 30 & 18th (18--22) \\
RELIG$\times$POLPARTY    & 30 & 18th (18--22) \\
RACE$\times$RELIG        & 30 & 18th (18--22) \\
RACE$\times$POLPARTY     & 28 & 23rd (23--24) \\
AGE$\times$POLPARTY      & 18 & 44th (44--46) \\
\bottomrule
\end{tabular}
\caption{Rank of each used pair among the 66 possible attribute pairs by
number of cells clearing $n\geq30$. Ranks are the best rank within each
tied block (competition ranking); the parenthesised range gives the
positions that block occupies.}
\label{tab:pair-rank}
\end{table}

\subsection{Prompt templates (verbatim)}
\label{app:templates}
For attribute type $t$ with human-readable attribute labels
$(\ell_1,\ell_2)$ and cell values $(v_1,v_2)$, the four templates are:

\begin{quote}\small
\textbf{T0 (third-person declarative).}
\texttt{This survey respondent's $\ell_1$ is $v_1$ and their $\ell_2$ is
$v_2$.}

\textbf{T1 (first-person).}
\texttt{I am answering this survey. My $\ell_1$ is $v_1$ and my $\ell_2$
is $v_2$.}

\textbf{T2 (structured profile).}
\texttt{Survey respondent profile:} \\
\texttt{- $\ell_1$: $v_1$} \\
\texttt{- $\ell_2$: $v_2$} \\
\texttt{This respondent is about to answer opinion questions.}

\textbf{T3 (QA format).}
\texttt{Question: What is this survey respondent's $\ell_1$ and $\ell_2$?}\\
\texttt{Answer: Their $\ell_1$ is $v_1$ and their $\ell_2$ is $v_2$.}
\end{quote}

The label pairs are: race/religion, race/political party affiliation,
race/political ideology, religion/political party affiliation, highest
level of education/household income, age group/political party
affiliation. \textsc{Tmean} averages the four extracted representations
(not the four prompts). Example instantiation of T0 for one cell:
\texttt{This survey respondent's age group is 30--49 and their political
party affiliation is Democrat.}

\section{Fidelity-Map Details}

\subsection{Anisotropy and mean-centering}
\label{app:anisotropy}
The last-token residual stream is strongly anisotropic: the ratio
$\lVert\overline{v}\rVert / \overline{\lVert v\rVert}$ over the 169 cell
embeddings is $0.950$ (it would be $\approx0$ for randomly oriented
vectors and $1$ if all vectors were identical in direction). At layer 17
the pairwise cosine distances have median $0.10$ and maximum $0.27$; the
\emph{same-type} pairs, which are the ones the task must separate, are
the tightest (maximum $0.12$).

Removing the shared direction ($v_i \leftarrow v_i - \overline{v}$) nearly
doubles the pooled correlation ($+0.185 \to +0.301$ at layer 17,
consistently $+0.07$ to $+0.16$ across all 32 layers), and the residual
variance is genuinely multidimensional afterwards (PC1 $31\%$, PC2
$20\%$, PC3 $12\%$). We nevertheless do \emph{not} adopt centering,
because the gain does not survive the decomposition that matters here:
within type---the only regime used for fidelity scoring and for the
$w_{ij}$ kernel---centering \emph{lowers} $\rrho$
(AGE$\times$POLPARTY $0.455\to0.375$; EDUCATION$\times$INCOME
$0.490\to0.454$; RACE$\times$RELIG $0.072\to0.033$). The pooled
improvement is an artifact of aligning scale \emph{between} types, not
sharpening signal \emph{within} them. Table~\ref{tab:centering}
summarizes this within-type comparison.

\label{app:centering}
\begin{table}[t]
\centering\small
\caption{Fidelity ($\rrho$ vs.\ survey truth) at layer 17 before and
after removing the shared direction ($v_i \leftarrow v_i - \bar{v}$).
Centering was run on three of the six types; full detail above.}
\label{tab:centering}
\begin{tabular}{lcccc}
\toprule
& Pooled & AGE$\times$POLPARTY & EDUCATION$\times$INCOME & RACE$\times$RELIG\\
\cmidrule(lr){2-2}\cmidrule(lr){3-5}
Read-out & \multicolumn{1}{c}{(mixes types)} & \multicolumn{3}{c}{within-type (the regime used throughout)}\\
\midrule
raw            & $+0.185$ & $+0.455$ & $+0.490$ & $+0.072$ \\
mean-centred   & $+0.301$ & $+0.375$ & $+0.454$ & $+0.033$ \\
\midrule
change & $\mathbf{+0.116}$ & $-0.080$ & $-0.036$ & $-0.039$ \\
\bottomrule
\end{tabular}
\end{table}

\subsection{Full fidelity map}
\label{app:map}
\begin{figure}[t]
\centering
\includegraphics[width=\linewidth]{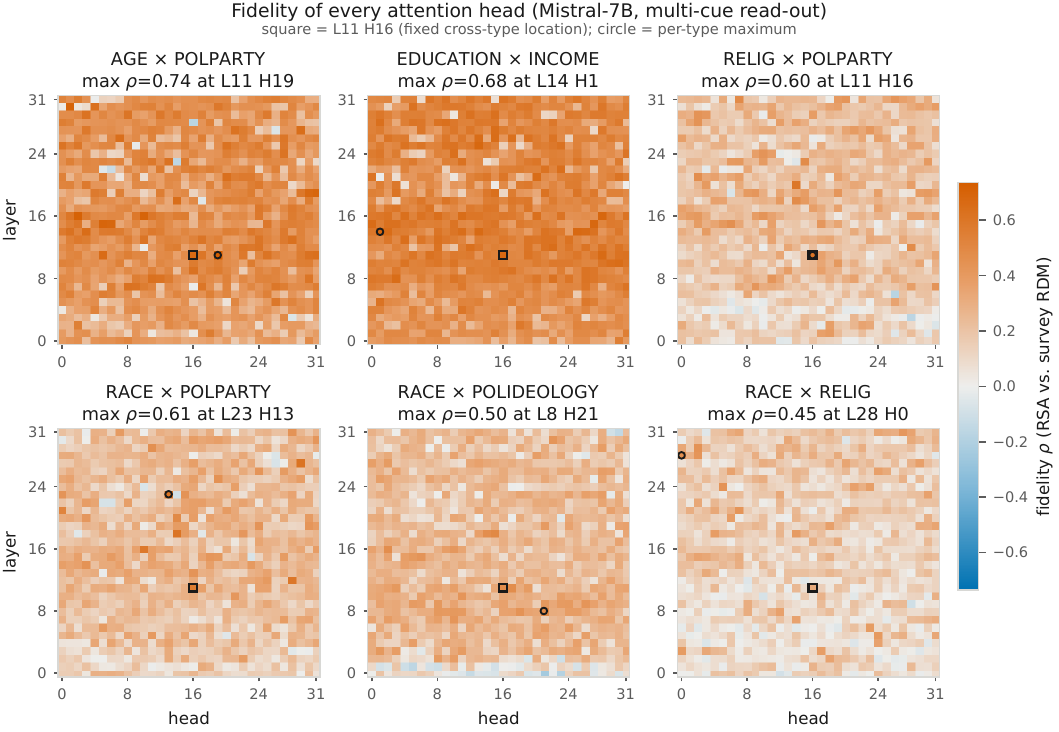}
\caption{The full fidelity map: every attention head (32 layers
$\times$ 32 heads) scored against survey ground truth, per attribute
type, multi-cue read-out. Square marks the fixed location \starhead;
circle marks that type's own maximum; the underlying table is described
above. In the two strongest types $84$--$92\%$ of the
1{,}024 heads exceed $\rrho=0.3$, against $5\%$ in RACE$\times$RELIG; the
level varies with depth rather than at random
(Table~\ref{tab:layerblock} above) and falls for race-containing types (bottom
row).
Maxima printed here are uncorrected and match the \emph{selected}
column of Table~\ref{tab:map}; see that table for held-out values.}
\label{fig:headmap}
\end{figure}

The table behind Figures~\ref{fig:headmap} and~\ref{fig:layercurve} (\texttt{peta\_kesetiaan\_full.csv}) holds
$32{,}670$ rows: $1{,}089$ locations $\times$ 5 template conditions
$\times$ 6 types. The layer-0 residual read-out is \texttt{NaN} by
construction---the final token is shared across cells, so that
representation is constant---which is why the residual curves start at
layer 1.

\label{app:quarter-table}
\begin{table}[t]
\centering\small
\caption{Fidelity averaged over all 32 heads within each quarter of the
stack (multi-cue \textsc{Tmean} read-out). Unlike
Figure~\ref{fig:layercurve}, which takes the best head per layer, a block
mean involves no selection, so these values are comparable across depths.
Bold marks each type's peak block.}
\label{tab:layerblock}
\begin{tabular}{lcccc}
\toprule
Type & L0--7 & L8--15 & L16--23 & L24--31\\
\midrule
AGE$\times$POLPARTY & 0.36 & $\mathbf{0.47}$ & 0.42 & 0.42 \\
EDUCATION$\times$INCOME & 0.43 & $\mathbf{0.53}$ & 0.46 & 0.48 \\
RELIG$\times$POLPARTY & 0.13 & 0.21 & $\mathbf{0.23}$ & 0.22 \\
RACE$\times$POLPARTY & 0.17 & $\mathbf{0.24}$ & 0.24 & 0.22 \\
RACE$\times$POLIDEOLOGY & 0.18 & $\mathbf{0.27}$ & 0.22 & 0.21 \\
RACE$\times$RELIG & 0.09 & 0.12 & 0.15 & $\mathbf{0.17}$ \\
\bottomrule
\end{tabular}
\end{table}

\label{app:headmap-full}

\subsection{Selection-correction details}
\label{app:selection}
All three corrections run on the extracted representations, no GPU
required (\path{notebooks/10_koreksi_seleksi_peta_local.ipynb}).

\textbf{Max-statistic permutation} (\texttt{cek2}, 2{,}000 permutations):
cell labels of the survey RDM are permuted, the whole $1{,}024$-head map
is recomputed, and the \emph{maximum} $\rrho$ is retained, giving the
distribution of what a champion selected from noise can reach. Null
means are $0.21$--$0.32$ and null 95th percentiles $0.34$--$0.46$; the
observed maxima exceed them in every type. This test runs on the
NaN-cleaned cell subset, on which the three race-type maxima are slightly
higher than the full-set values in Table~\ref{tab:map}
($0.45\!\to\!0.49$, $0.50\!\to\!0.52$, $0.61\!\to\!0.65$); null and
observation are computed on the same subset, so the comparison is
internally consistent.

\textbf{Held-out template} (\texttt{cek1}): the head is selected on three
templates and scored on the fourth, rotating over all four folds.

\textbf{Split-half} (\texttt{cek3}, 200 resamples): cells are split at
random, the head is selected on one half and scored on the other; we
report the median. The same table records how often each head is
re-selected---\starhead{} in 87/200 splits for RELIG$\times$POLPARTY
versus a maximum of 22/200 for any head in RACE$\times$RELIG, the
quantitative form of the ``head choice is a lottery'' statement in
Sec.~\ref{sec:act2}.

\textbf{Fixed-location test} (\texttt{cek4}): \starhead{} and L18\,H14
are fixed in advance and scored per type with a per-location permutation
test (1{,}000 permutations) and a conservative $\times1024$ Bonferroni
threshold. \starhead{} passes in all six types; L18\,H14 fails only in
RACE$\times$RELIG ($p_{\text{raw}}=0.004$).

\subsection{Random-projection control for the head--residual comparison}
\label{app:randproj}
The three candidate families (1{,}024 heads; 33 residual layers; 1{,}024
random 128-dimensional Gaussian projections of the residual stream, 32
per layer, fixed across folds) are put through one identical
held-out-template selection pipeline. Held-out fidelity per type
(head / residual / projection): AGE $0.52/0.39/0.42$; EDU
$0.54/0.51/0.49$; RELIG$\times$PP $0.44/0.17/0.18$; RACE$\times$PP
$0.47/0.28/0.30$; RACE$\times$PI $0.38/0.26/0.29$; RACE$\times$RELIG
$0.29/0.16/0.17$. (This pipeline's held-out variant selects on the mean
of three templates and is slightly more conservative than the one behind
Table~\ref{tab:map}; comparisons are valid within this table only.)

\subsection{Lexical-baseline control (partial correlations)}
\label{app:lexical}
The lexical RDM embeds either the two attribute-value phrases (averaged)
or the full T0 identity sentence with all-MiniLM-L6-v2 \citep{minilm2020}
and takes cosine distances. Partial Spearman is computed on ranks;
significance is by cell-label permutation of the survey RDM (2{,}000
permutations), holding both model and lexical RDMs fixed. Per type
(value-phrase variant): lexical-vs-survey $\rrho$ = AGE $0.39$, EDU
$0.64$, RELIG$\times$PP $0.21$, RACE$\times$PP $0.21$, RACE$\times$PI
$0.31$, RACE$\times$RELIG $0.28$; \starhead's partial fidelity = $0.59$,
$0.35$, $0.59$, $0.58$, $0.37$ (all $p<0.0005$) and $0.23$ ($p=0.056$)
respectively. The full-sentence variant gives the same qualitative
picture with higher lexical baselines (up to $0.64$ for AGE).

\subsection{Reliability ceiling for the fidelity map}
\label{app:reliability}
Survey-side reliability simulates two independent respondent samples per
(cell, question) at the real unweighted sample sizes and correlates the
two resulting RDMs (10 repetitions): $0.985$--$0.994$ per type.
Model-side reliability correlates RDMs built from disjoint template
halves (all three 2v2 splits, averaged): $0.72$--$0.94$ at \starhead.
The attenuation ceiling $\sqrt{r_{\text{survey}}\, r_{\text{model}}}$ is
$0.84$--$0.96$; observed fidelity reaches $39$--$74\%$ of it.

\subsection{Sensitivity of the survey-side ruler}
\label{app:dreal-sens}
Normalising each question's WD by its scale range changes \starhead's
fidelity by at most $0.03$ in any type (e.g., AGE $0.68\to0.69$; EDU
$0.66\to0.63$). Restricting to questions answered by \emph{every} cell of
a type is possible for three types (77--102 common questions): fidelity
holds for AGE ($0.68\to0.64$) and RELIG$\times$POLPARTY ($0.61\to0.56$;
best head $0.72$), and drops for EDUCATION$\times$INCOME
($0.66\to0.44$). For the three race-containing types no question is
answered by all cells, so this variant is undefined there.

\section{Causal-Experiment Details}

\subsection{Patching protocol details}
\label{app:patching}
\textbf{Read/write point.} All interventions are applied to the input of
\texttt{self\_attn.o\_proj}---the per-head activations before the
out-projection mixes them---so a head is addressed as a $128$-dimensional
slice of the $4{,}096$-dimensional vector. The patch is an interpolation
toward the donor,
\begin{equation}
x \;\leftarrow\; x + \alpha\,(d - x),
\end{equation}
with $\alpha=1$ (full replacement) and $\alpha=5$ (amplification) as the
two reported settings, $d$ the donor group's activation at the
corresponding position.

\textbf{v1 (Sec.~\ref{sec:gap}, weak instrument).} Identity is a natural
sentence, the patch is applied at the final (prediction) token only, and
1--3 heads are patched at natural scale.

\textbf{v2 (Sec.~\ref{sec:v2}, strong instrument).} Identity is localized
to the answer token of a QA-formatted demographic block
(\texttt{Question: What is this survey respondent's \dots{} / A. \dots{} /
Answer: B}), and the patch is applied at those identity-answer token
positions, so the edit propagates to every later token. Conditions:
\texttt{patch\_L11\_all32} (all 32 heads at layer 11),
\texttt{patch\_L11H16} (the star head alone),
\texttt{patch\_L11L18\_all} (both layers), \texttt{patch\_L11\_all32\_x5}
($\alpha=5$), and \texttt{patch\_randhead} (a single randomly drawn head,
excluded from $\{$L11\,H16, L18\,H14, L11\,H19$\}$, fixed for the run) as
the control. A \texttt{self\_patch} sanity condition---patching a cell
with its own activations, which must produce zero shift---is run on the
first three pairs of each type. Scope: 3 types $\times$ 12 pairs $\times$
20 questions, the questions chosen per pair by largest real inter-group
Wasserstein distance, giving $n=240$ per type--condition cell.

\textbf{Sweep (Sec.~\ref{sec:sweep}).} The v2 all-32-heads intervention is
repeated at every layer, for all six attribute types (three in the first
sweep, three added later with identical settings: 12 pairs $\times$ 20
maximum-disagreement questions, $n=240$ items). The selection correction
is a sign-flip permutation test (2{,}000 permutations) in which the same
random sign is applied to an item across all layers, preserving the
between-layer correlation structure; the statistic is the maximum $t$ over
the 32 layers. All six types were corrected in one run with one seed;
re-running that code on the first three types reproduces the earlier
figures to within permutation noise (e.g.\ $t=4.098$ vs.\ $4.10$ for
RACE$\times$RELIG at layer 1).

\textbf{Read-out.} The predicted opinion distribution is the softmax over
the first $n_{\text{opt}}$ answer-letter token ids at the position after
\texttt{Answer:}, and is compared to the target group's real distribution
by Wasserstein distance on the question's ordinal scale. The reported
shift is
$\wdist(\hat{p}_A, p^{\mathrm{real}}_B) - \wdist(\hat{p}_{\text{patched}},
p^{\mathrm{real}}_B)$: positive means the patch moved the prediction
toward the donor group's \emph{truth}, not merely toward the model's own
prediction for that group.

\textbf{Donor controls} (for the \starhead{} backfire,
Sec.~\ref{sec:v2}). On the same 12 pairs and 20 questions per type
(RACE$\times$RELIG and, as a negative control, AGE$\times$POLPARTY), the
single-head patch is repeated under four donors captured at the same
positions: (i) the paired group's activation (replication of the original
condition); (ii--iii) the activation of two other same-type identities,
drawn at random per pair excluding both members; and (iv) a
dimension-shuffled donor---the paired group's vector with its 128 head
dimensions permuted by a fixed per-pair permutation, matching location,
norm, and coordinate statistics while destroying structure. Donors are
captured per (cell, question) exactly as in the main experiment. All
inference is pair-level (exact sign-flip, two-sided). Raw output movement
is tracked as $\wdist(\hat{p}_{\text{patched}}, \hat{p}_A)$ so that
harm can be compared against perturbation size across conditions.

\subsection{Cluster-robust causal inference}
\label{app:cluster}
For every test we first average within pair (12 values), then enumerate
all $2^{12}=4{,}096$ sign-flip combinations exactly; max-statistic
variants take the maximum $t$ over the 32 layers within each flip.

\subsection{Identity-swap ceiling: instruction-tuned model and per-type spread}
\label{app:ceiling-instruct}
\paragraph{Instruction-tuned model.}
We re-measured
the identity-swap ceiling on Mistral-7B-Instruct-v0.2 under its own chat
template, with the same cells and questions (40 per type, all same-type
cell pairs; under this estimator the base model's swap movement is
$3.2\%$ of its total error, rather than the ${<}2\%$ above, which is
computed on the original question set). The near-invariance replicates:
the instruct model's full-identity swap moves predictions by $3.9\%$ of
its total error. What chat tuning changes is accuracy and direction, not
identity-dependence: absolute error nearly doubles (mean $\wdist$
$0.66$ vs.\ $0.37$), while the small movement becomes reliably
directionally correct (group $A$'s prediction closer to $A$'s own truth
than to $B$'s in $64\%$ of pairs; sign test $p<0.001$ in all six types,
vs.\ three of six at $p<0.05$, one at $p<0.001$, for the base model). An
instruction-tuned model is thus \emph{on average} slightly more
identity-responsive and substantially less accurate; the effect is not
uniform, and one type moves the other way
(RACE$\times$POLPARTY, $3.6\%\!\to\!1.5\%$; Table~\ref{tab:ceiling}). The
near-invariance itself is a property of the model family, not of the
raw-completion interface, though both estimates assume the
letter-probability read-out (Sec.~\ref{sec:limitations}).

\begin{table}[t]
\centering\small
\setlength{\tabcolsep}{5pt}
\caption{Identity-swap ceiling per attribute type, base vs.\
instruction-tuned Mistral (40 questions per type, all same-type cell
pairs). \emph{Swap} = movement of the predicted distribution under a full
identity replacement, as a percentage of that type's total prediction
error. $\wdist$ = mean absolute error to survey truth (lower is better).
\emph{Dir.} = fraction of pairs where group $A$'s prediction is closer to
$A$'s own truth than to $B$'s, with the sign-test $p$; $0.50$ is a coin
flip.}
\label{tab:ceiling}
\begin{tabular}{lccccccc}
\toprule
& \multicolumn{2}{c}{Swap (\% of error)} & \multicolumn{2}{c}{Mean $\wdist$}
& \multicolumn{2}{c}{Dir.\ ($p$)}\\
\cmidrule(lr){2-3}\cmidrule(lr){4-5}\cmidrule(lr){6-7}
Type & base & instr. & base & instr. & base & instr.\\
\midrule
AGE$\times$POLPARTY & 2.6 & 3.4 & 0.36 & 0.65 & 0.56\,(.045) & 0.75\,($<$.001) \\
EDUCATION$\times$INCOME & 1.0 & 2.4 & 0.33 & 0.59 & 0.55\,(.002) & 0.62\,($<$.001) \\
RELIG$\times$POLPARTY & 1.6 & 3.4 & 0.37 & 0.64 & 0.52\,(.293) & 0.65\,($<$.001) \\
RACE$\times$POLPARTY & 3.6 & 1.5 & 0.35 & 0.70 & 0.54\,(.061) & 0.59\,($<$.001) \\
RACE$\times$POLIDEOLOGY & 9.2 & 9.9 & 0.37 & 0.66 & 0.62\,($<$.001) & 0.64\,($<$.001) \\
RACE$\times$RELIG & 1.2 & 2.9 & 0.46 & 0.70 & 0.52\,(.297) & 0.59\,($<$.001) \\
\bottomrule
\end{tabular}
\end{table}

\paragraph{Per-type spread.}
\label{app:ceiling-spread}
Per type (Table~\ref{tab:ceiling}), the identity swap moves
RACE$\times$POLIDEOLOGY by $9.2\%$ of its error but
EDUCATION$\times$INCOME by only $1.0\%$. First, the ranking is not the
fidelity ranking: the two extremes
of \emph{fidelity}---RELIG$\times$POLPARTY, among the most faithful types
(Table~\ref{tab:map}), and RACE$\times$POLIDEOLOGY, mid-table---sit at
$1.6\%$ and $9.2\%$ respectively, i.e.\ the more faithful type is the
\emph{less} identity-responsive one. The same inversion holds under the
independent ceiling estimator of Table~\ref{tab:sweep}, computed on the
causal item set: RELIG$\times$POLPARTY is last of six there ($0.0020$)
and RACE$\times$POLIDEOLOGY second ($0.0176$). Second, the two estimators
do \emph{not} agree in general---EDUCATION$\times$INCOME is last here and
first there ($0.0194$), reflecting different question sets and absolute
vs.\ error-normalised scaling. The spread therefore supports the
fidelity--use inversion at the two ends, but not a per-type ordering.

\subsection{Amplification and donor controls}
\label{app:amplification}
For RACE$\times$RELIG, replacing
128 dimensions mid-computation is a destructive intervention, so
we ran a donor-control experiment on the same pairs and questions
(independent run; the effect itself replicates, pair-level
$t=-2.4$, exact $p=0.005$).
A \emph{dimension-shuffled} donor---the target group's own
vector with its 128 head dimensions permuted, so the location,
norm, and coordinate statistics match while the structure is
destroyed---perturbs the output \emph{twice} as strongly in raw
movement, yet produces no systematic drift at all (pair-level
$t=-0.1$): if harm merely scaled with disruption, this condition
would be the most harmful, and it is the least. The consistent
away-from-truth drift appears only with a \emph{coherent identity
donor}. Whether it further requires the \emph{paired} group's
donor is suggestive but unsettled: donors from other same-type
identities trend the same way weakly (each n.s.), and the direct
paired contrast against their mean reaches $t=-2.0$, exact
$p=0.073$. Narrowly, then: single-unit intervention at an
apparently faithful location remains unreliable in direction.

Scaling the patch by $5\times$
(matching the amplification levels used in SAE-feature steering,
\citealp{llmopinions2026}) makes the shift \emph{negative} in all three
types (condition not shown in Table~\ref{tab:patchv2}; full results in
Appendix~\ref{app:patching}). Linear extrapolation in raw activation space likely
pushes representations off-manifold, unlike steering in a constrained SAE
feature space---amplifying a raw signal is not a free strengthening
operation.

\subsection{Layer-sweep procedure, control sensitivity, and caveats}
\label{app:sweep-procedure}
\paragraph{Procedure.}
For each pair--question item we assemble a single
batch of 33 copies of the prompt: one patched at each of the 32 layers,
plus one patched at a randomly chosen layer as a control, so all
conditions see identical inputs and one forward pass answers the whole
sweep ($n{=}240$ pair--question items per type). The statistic per layer
is the paired $t$ of its directional shift against the random-layer
control. Two selection problems require correction. First, the winner is
selected from 32 candidate layers, so we apply a max-statistic permutation
correction \citep{nichols2002}, with the same sign flip applied across all
layers to preserve the between-layer correlation structure. Second, and
more consequentially, the 240 items are \emph{clustered} into 12 pairs
(Sec.~\ref{sec:setup-selection}), so all primary $p$-values below flip
signs at the \emph{pair} level: per-pair mean shifts, all $2^{12}$ flip
combinations enumerated exactly. Item-level statistics are shown as
descriptive detail only---treating the 240 items as independent
understates the $p$-values several-fold.

One further wrinkle affects the layer-1 reading specifically: layer~1
was noticed as a candidate locus only during the first sweep, which
included RACE$\times$RELIG, so for that one type the layer-1 column of
Table~\ref{tab:fixedlayer} is not selection-free (its pair-level
$p=0.016$ should be read with that caveat, and the ``early-layer hub''
reading stays a hypothesis for the cross-model replication); for the
three types added later, both the L11 and L1 columns are genuinely
fixed in advance.

\paragraph{Control sensitivity.}
\label{app:sweep-control-sens}
The random-layer control can itself land
on a causally relevant layer, deflating measured effects; and
``shift'' is defined against the unpatched baseline prediction, so a
no-patch reference exists implicitly. Re-running every pair-level test
against zero instead of against the control changes no conclusion (e.g.,
RACE$\times$POLIDEOLOGY@L11: $t=3.24$, $p=0.0022$ vs.\ zero; full table
in the released analysis code).

\paragraph{Five caveats.}
\label{app:sweep-caveats}
(i) Absolute effects are small
throughout ($0.001$--$0.012$), commensurate with output ceilings that are
themselves small; the surviving effect is statistically robust, not
practically large. (ii) Most loci were \emph{discovered} by the sweep
rather than predicted, so they warrant replication in a second model
(Sec.~\ref{sec:robustness})---though the L11 column of
Table~\ref{tab:fixedlayer} is free of per-type selection. (iii) With six
types, any cross-type correlation is weakly powered; our claim rests on
the two counterexamples, not on the correlation coefficient. (iv) What
layer 1 encodes mechanistically remains open: a check on already-collected
residual streams was inconclusive because the layer-0 read-out is
degenerate in our prompt design (the final token is shared across cells),
so we cannot contrast layer 1 against pure embeddings; layers 1 and 2 are
near-identical in geometry ($\rrho=0.90$) and both differ appreciably from
layer 11 ($\rrho=0.78$); see Sec.~\ref{sec:limitations}. (v) A
single-layer patch may fail to detect a causally-used representation that
is redundantly encoded across many layers---the pattern a high,
broadly-distributed fidelity (Figure~\ref{fig:headmap}) would predict. Our one direct check argues
against strong compensation, though only for one type: AGE$\times$POLPARTY's
two-layer patch (L11+L18 jointly, Sec.~\ref{sec:v2}) moves the
pair-level effect only marginally beyond the single-layer result
($p=0.019\to0.016$). EDUCATION$\times$INCOME---the type this caveat
matters most for---has not been tested under a multi-layer patch;
redundant, single-layer-invisible use remains an alternative we have
not ruled out for it.

\paragraph{Magnitude versus consistency.}
\label{app:magnitude-consistency}
Three separate times
in this sweep the layer with the largest raw shift fails correction while
a smaller, steadier layer passes: RELIG$\times$POLPARTY (L12/L11 largest,
L8 passes), RACE$\times$RELIG (L7 $+0.0030$ and L10 $+0.0026$ fail, L1
passes at $+0.0010$, with the selection-inflation caveat given above for
this type), and EDUCATION$\times$INCOME (largest shift in the
paper, nothing passes at either clustering level). Reporting mean effect
sizes without a selection-and-cluster-corrected variance test would have
produced three wrong localisations here.

\subsection{The fidelity map in the QA context}
\label{app:qa-context-map}
Sec.~\ref{sec:act2}'s fidelity
is scored on an identity-only prompt (no opinion question, last token),
whereas the causal and probe experiments are scored in a full QA
context at the opinion-answer position. A tempting alternative reading
of the dissociation is therefore not ``the model does not consult its
map'' but ``the faithful geometry is not the one that exists once a
question is present.'' We test this on the activations already
collected for the probe (Sec.~\ref{sec:probe}), holding question
composition fixed across cells---each cell is averaged over the same
question set, the questions answered by every cell of its type, since
coverage is only ${\ge}60\%$ per cell and unmatched subsets would
themselves contribute inter-cell distance. The resulting RDM is
near-perfectly reliable (split-half $\rrho=0.98$--$0.99$ in the five
types with enough all-cell-shared questions to split, above the
identity-only side's $0.72$--$0.94$), so what follows is a property of
the representation, not of measurement noise.

\textbf{The map does degrade}: in five of six types QA-context fidelity falls well below the
identity-only value (EDUCATION$\times$INCOME $0.67\to0.10$;
RACE$\times$POLPARTY $0.59\to0.17$; RELIG$\times$POLPARTY
$0.60\to0.23$; RACE$\times$POLIDEOLOGY $0.47\to0.24$;
AGE$\times$POLPARTY $0.67\to0.39$), the exception being
RACE$\times$RELIG, which \emph{rises} ($0.33\to0.49$).
Sec.~\ref{sec:act2} therefore characterises the geometry of an
identity-only prompt, and only a weaker version of that geometry
survives into the context where the model actually answers.
\textbf{The dissociation, however, does not depend on the ruler}: with
the QA-context ruler at the a-priori locus L11 the correlation is
$\rrho=+0.03$ ($p=0.96$), the same null the identity-only ruler gives.
The type with the clearest causal locus (RACE$\times$POLIDEOLOGY) is
mid-pack on the QA ruler too, and the most faithful type under it
(RACE$\times$RELIG) has no detectable L11 effect ($p=0.27$). At each
type's own \emph{selected} best layer the correlation is ruler-unstable
($-0.77$ identity-only vs.\ $+0.71$ QA-context, both n.s.), which is
the same caution we already attach to that quantity. Measured in the
very context where the causal experiments look for it, fidelity still
fails to predict causal use---so ``we measured the wrong map'' does not
explain the dissociation away.

\section{Probe Details}

\subsection{Probe protocol and the group-ordering test}
\label{app:probe}
\textbf{Data.} Questions are selected for \emph{coverage} rather than
disagreement: within each attribute type we keep questions answered by at
least $60\%$ of that type's cells (minimum 10), then draw a random sample
of 250 of them (seed 42). Random sampling matters: the highest-coverage
questions concentrate in two or three survey waves, which would restrict
the probe to a narrow topic range. The resulting set is $33{,}395$
(cell, question) rows, $4{,}051$--$7{,}309$ per type, spanning up to all
15 ATP waves per type. Activations are captured at the opinion-answer
position and stored in half precision.

\textbf{Fitting.} Features are reduced with PCA (128 components, fit on
training rows only) and a ridge regression predicts the six-dimensional
padded response distribution; predictions are truncated to the question's
option count and renormalised. Evaluation is leave-one-cell-out: every
prediction is for a demographic cell absent from training. The ridge
penalty is chosen inside each training fold by a further split over cells
(median selected value $\alpha=1$). Scores are mean Wasserstein distance
to the real distribution on the question's ordinal scale.

\textbf{References.} ``Mouth'' is the softmax over answer-letter tokens
from the same forward pass. The group-blind baseline predicts, for each
held-out row, the mean real distribution of the \emph{other} cells on that
question (training cells only).

\textbf{Group-ordering test.} Correlations are Spearman. The
leave-one-out artifact the half-split design of Sec.~\ref{sec:probe}
avoids is severe: leave-one-out predictions are anti-correlated with the
held-out value by construction, driving the group-blind baseline to
$\rrho\approx-0.99$.

\textbf{Reliability ceiling.} To calibrate the group-ordering numbers we
simulate two independent survey samples per cell (multinomial draws at
that cell's own unweighted sample size) and run the same test between
them. This gives $+0.81$ to $+0.86$ per type, i.e.\ the maximum a perfect
read-out could achieve given survey sampling noise alone.

\subsection{Probe fitting procedure}
\label{app:probe-procedure}
For every (cell, question) pair we collect the
activation at the opinion-answer position and fit a ridge probe (128 PCA
components) to predict the group's \emph{real} response distribution
(full protocol in Appendix~\ref{app:probe}).
Evaluation is leave-one-cell-out: the probe is always scored on a
demographic cell it never saw, and the regularisation strength is chosen
inside the training fold. We read three locations---the full layer 11
($4{,}096$ dims), the single head \starhead{} ($128$ dims), and layer 1
($4{,}096$ dims)---and compare against two references: the model's own
answer distribution (``the mouth''), and a group-blind baseline that
predicts the mean real distribution of the \emph{other} cells on that
question. Broad-coverage questions are used throughout ($250$ questions
per type, each answered by at least $60\%$ of that type's cells;
$33{,}395$ rows in total), which is what the earlier, narrower question
set could not support.

\section{The Failed Consistency-Loss Pilot}
\label{app:lgroup}
Sec.~\ref{sec:act1} is motivated by a downstream failure, reported
here for
completeness. We had implemented a group-consistency auxiliary loss
$L_{\text{group}}$ that pulls the model's predictions for demographically
similar cells together, weighting pairs by a kernel $w_{ij}$ built from
last-token residual-stream embeddings. The pilot failed; the diagnosis:

\begin{enumerate}
  \item $L_{\text{group}}$ performed no better than a plain probe, and
        both lost to simple shrinkage; it was insensitive to $N$.
  \item The embeddings of all 26 RACE$\times$RELIG cells occupied one
        narrow cone (all pairwise similarities $>0.91$).
  \item A \emph{random} probe already drove $L_{\text{group}}$ to
        $\approx0$---the failure was geometric, not a training problem.
  \item The highest-weighted neighbours under $w_{ij}$ were frequently
        not the cells with the most similar real answers.
  \item Raising $\lambda$ from 1 to 100 monotonically worsened accuracy.
  \item A single shared probe let 23 data-rich cells dominate the sparse
        target cell.
\end{enumerate}

Points 2--4 are the direct motivation for asking where a \emph{better}
read-out might live, which is this paper's question.

\end{document}